\newif\ifarxiv
\arxivtrue   % set \arxivfalse for the ACM version
\documentclass[sigconf]{acmart}
\setcopyright{none}

\renewcommand\footnotetextcopyrightpermission[1]{} % removes the permission footnote
\AtBeginDocument{%
  }

\setcopyright{none}
\usepackage{enumitem}
\usepackage{multirow}
\usepackage{booktabs}
\usepackage{caption}
\usepackage{array}
\usepackage{makecell}
\usepackage{color}
\usepackage{xcolor}

\usepackage{subcaption}
\usepackage{threeparttable}
\usepackage{pifont}
\usepackage{pdfpages}
\usepackage{pdfcomment}
\usepackage{tikz}
\usepackage{pifont}
\usepackage{float}
\usepackage{bm}
\usepackage{listings} % 代码展示
\usepackage{xcolor} % 使用颜色
\usepackage{minted}
\usepackage{placeins}
\usepackage[ruled,vlined]{algorithm2e}

\DeclareMathOperator*{\argmax}{arg\,max}
\DeclareMathOperator*{\argmin}{arg\,min}

\begin{document}

% \title{A Bayesian Adversarial Approach to Robust Trading with a Controllable Financial Simulator}

\title{Bayesian Robust Financial Trading with \\ Adversarial Synthetic Market Data}
\author{Haochong Xia}
\authornote{Both authors contributed equally to this research.}
% \author{Simin Li}
% \email{lisiminsimon@buaa.edu.cn}
\affiliation{%
  \institution{Nanyang Technological University}
  % \city{Singapore}
  \country{Singapore}
}
\email{haochong001@e.ntu.edu.sg}
\orcid{0009-0004-2947-5947}

\author{Simin Li}
\authornotemark[1]
\affiliation{%
  \institution{Beihang University}
  \city{Beijing}
  \country{China}}
\email{lisiminsimon@buaa.edu.cn}

\author{Ruixiao Xu}
\affiliation{%
  \institution{Beihang University}
  \city{Beijing}
  \country{China}}
\email{xuruixiao@buaa.edu.cn}

\author{Zhixia Zhang}
\author{Hongxiang Wang}
\author{Zhiqian Liu}
\affiliation{%
  \institution{Beihang University}
  \city{Beijing}
  \country{China}}
\email{22376220@buaa.edu.cn}
\email{23371110@buaa.edu.cn}
\email{22373024@buaa.edu.cn}

% \author{Hongxiang Wang}
% \affiliation{%
%   \institution{Beihang University}
%   \city{Beijing}
% \email{23371110@buaa.edu.cn}

% \author{Zhiqian Liu}
% \affiliation{%
%   \institution{Beihang University}
%   \city{Beijing}
% \email{22373024@buaa.edu.cn}

\author{Teng Yao Long}
\author{Molei Qin}
\author{Chuqiao Zong}
\affiliation{%
  \institution{Nanyang Technological University}
  % \city{Singapore}
  \country{Singapore}
}
\email{yaolong001@e.ntu.edu.sg}
\email{molei001@e.ntu.edu.sg}
\email{ZONG0005@e.ntu.edu.sg}

% \author{Molei Qin}
% \affiliation{%
%   \institution{Nanyang Technological University}
%   % \city{Singapore}
%   % \state{Ohio}
%   \country{Singapore}
% }
% \email{molei001@e.ntu.edu.sg}

% \author{Chuqiao Zong}
% \affiliation{%
%   \institution{Nanyang Technological University}
%   % \city{Singapore}
%   % \state{Ohio}
%   \country{Singapore}
% }
% \email{ZONG0005@e.ntu.edu.sg}

\author{Bo An}
\authornote{Corresponding Author.}
\affiliation{%
  \institution{Nanyang Technological University}
  % \city{Singapore}
  \country{Singapore}
}
\email{boan@ntu.edu.sg}

\renewcommand{\shortauthors}{Haochong Xia et al.}
% \authornotemark[2]

\begin{abstract}

Algorithmic trading relies on machine learning models to make trading decisions. Despite strong in-sample performance, these models often degrade when confronted with evolving real-world market regimes, which can shift dramatically due to macroeconomic changes—e.g., monetary policy updates or unanticipated fluctuations in participant behavior. We identify two core challenges that perpetuate this mismatch: (1) insufficient robustness in existing policy against uncertainties in high-level market fluctuations, and (2) the absence of a realistic and diverse simulation environment for training, leading to policy overfitting. To address these issues, we propose a Bayesian Robust Framework that systematically integrates a macro-conditioned generative model with robust policy learning. On the data side, to generate realistic and diverse data, we propose a macro-conditioned GAN-based generator that leverages macroeconomic indicators as primary control variables, synthesizing data with faithful temporal, cross-instrument, and macro correlations. On the policy side, to learn robust policy against market fluctuations, we cast the trading process as a two-player zero-sum Bayesian Markov game, wherein an adversarial agent simulates shifting regimes by perturbing macroeconomic indicators in the macro-conditioned generator, while the trading agent—guided by a quantile belief network—maintains and updates its belief over hidden market states. The trading agent seeks a Robust Perfect Bayesian Equilibrium via Bayesian neural fictitious self-play, stabilizing learning under adversarial market perturbations. Extensive experiments on 9 financial instruments demonstrate that our framework outperforms 9 state-of-the-art baselines. In extreme events like the COVID pandemic, our method shows improved profitability and risk management, offering a reliable solution for trading under uncertain and rapidly shifting market dynamics.

\end{abstract}



% \setcopyright{acmlicensed}
% \copyrightyear{2025}
% \acmYear{2025}
% \acmConference[KDD '25]{KDD '25: Proceedings of the 31th ACM SIGKDD Conference on Knowledge Discovery and Data Mining}{August 3-7, 2025}{Toronto, Canada}
% \acmBooktitle{Proceedings of the 31th ACM SIGKDD Conference on Knowledge Discovery and Data Mining, August 3--7, 2025, Toronto, Canada}
% % \copyrightyear{2018}
% % \acmYear{2018}
% % \acmDOI{XXXXXXX.XXXXXXX}

% \renewcommand\footnotetextcopyrightpermission[1]{} % removes footnote with conference information in first column  

% % remove the copyright information
% \settopmatter{printacmref=false} % Removes citation information below abstract
% \renewcommand\footnotetextcopyrightpermission[1]

% \input{converletter}

\twocolumn
\setcounter{section}{0}
\maketitle
\newcommand\kddavailabilityurl{https://doi.org/10.5281/zenodo.18144122}
\ifdefempty{\kddavailabilityurl}{}{
\begingroup\small\noindent\raggedright\textbf{Resource Availability:}\\
% please change the following context to include multiple artifacts if necessary, including data, models, code, etc.
The source code of this paper has been made publicly available at \url{https://github.com/XiaHaochong98/Bayesian-Robust-Financial-Trading-with-Adversarial-Synthetic-Market-Data}.
\endgroup
}

\section{INTRODUCTION}

\begin{figure}[!t]
    \centering
    \begin{subfigure}{0.135\textwidth}
        \centering
        \includegraphics[width=\linewidth]{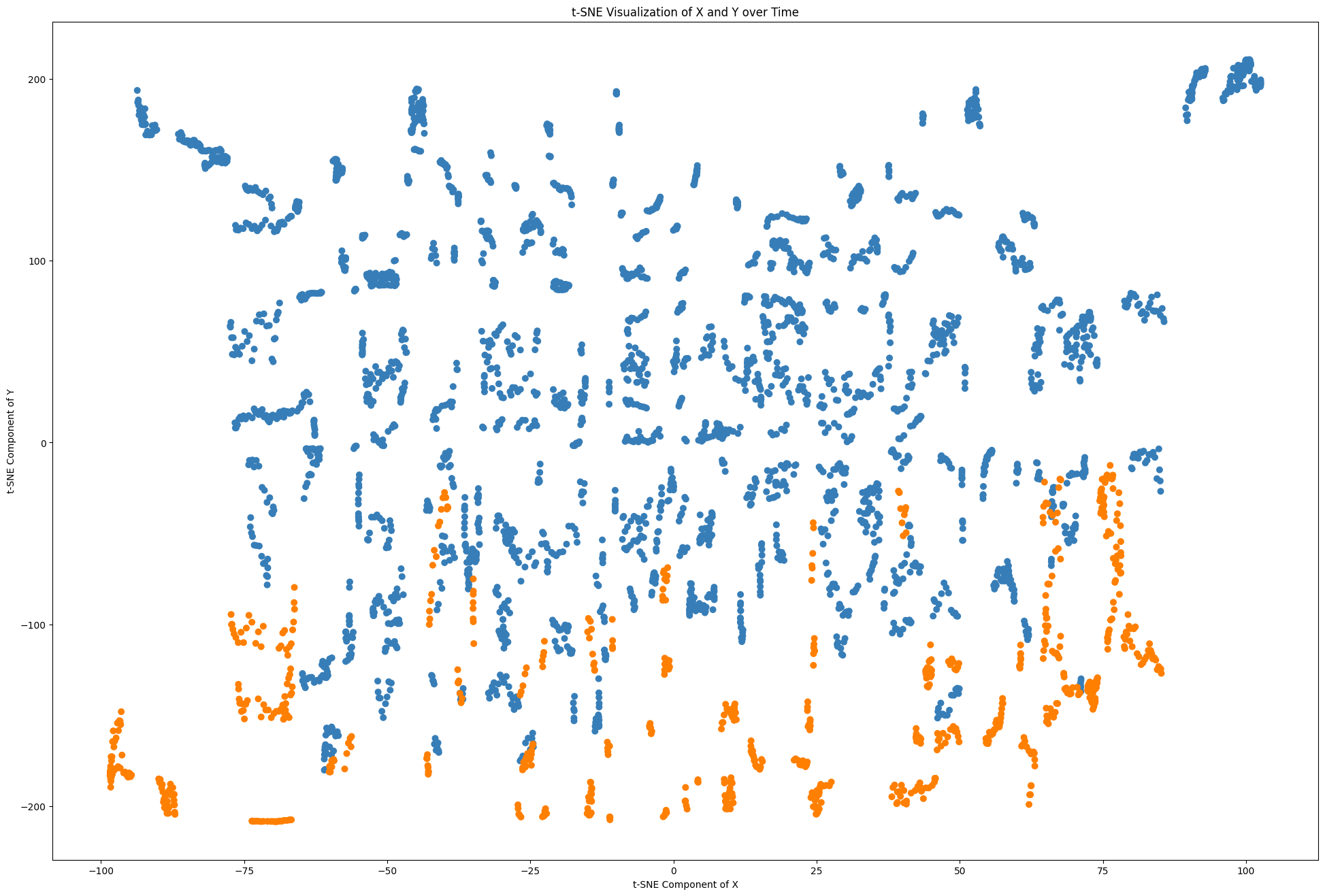}
        \captionsetup{font=small}
        \caption{UNG}
        \label{fig:sub1}
    \end{subfigure}%
    \hspace{0.01\textwidth}
    \begin{subfigure}{0.135\textwidth}
        \centering
        \includegraphics[width=\linewidth]{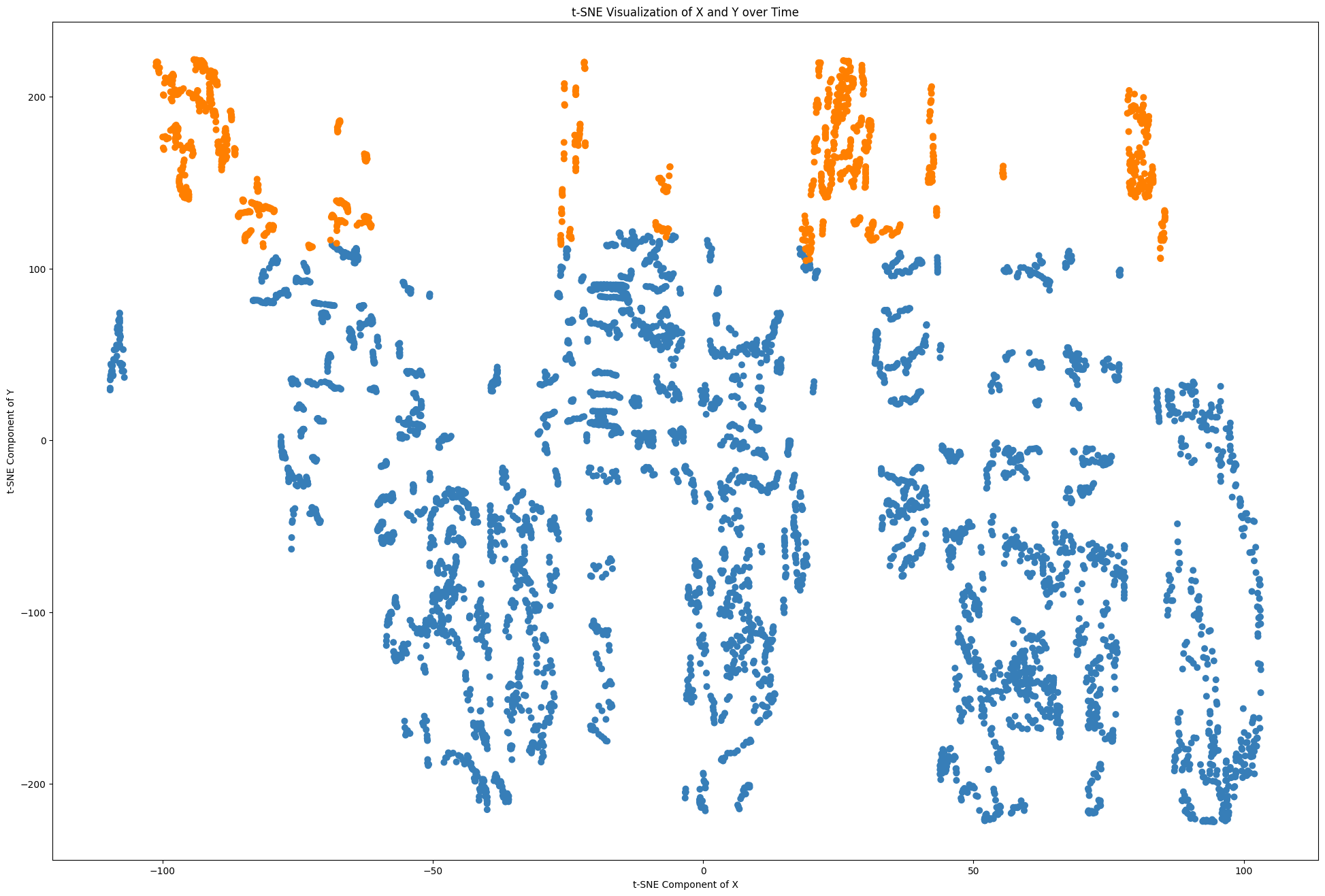}
        \captionsetup{font=small}
        \caption{QQQ}
        \label{fig:sub2}
    \end{subfigure}%
    \hspace{0.02\textwidth}
    \begin{subfigure}{0.135\textwidth}
        \centering
        \includegraphics[width=\linewidth]{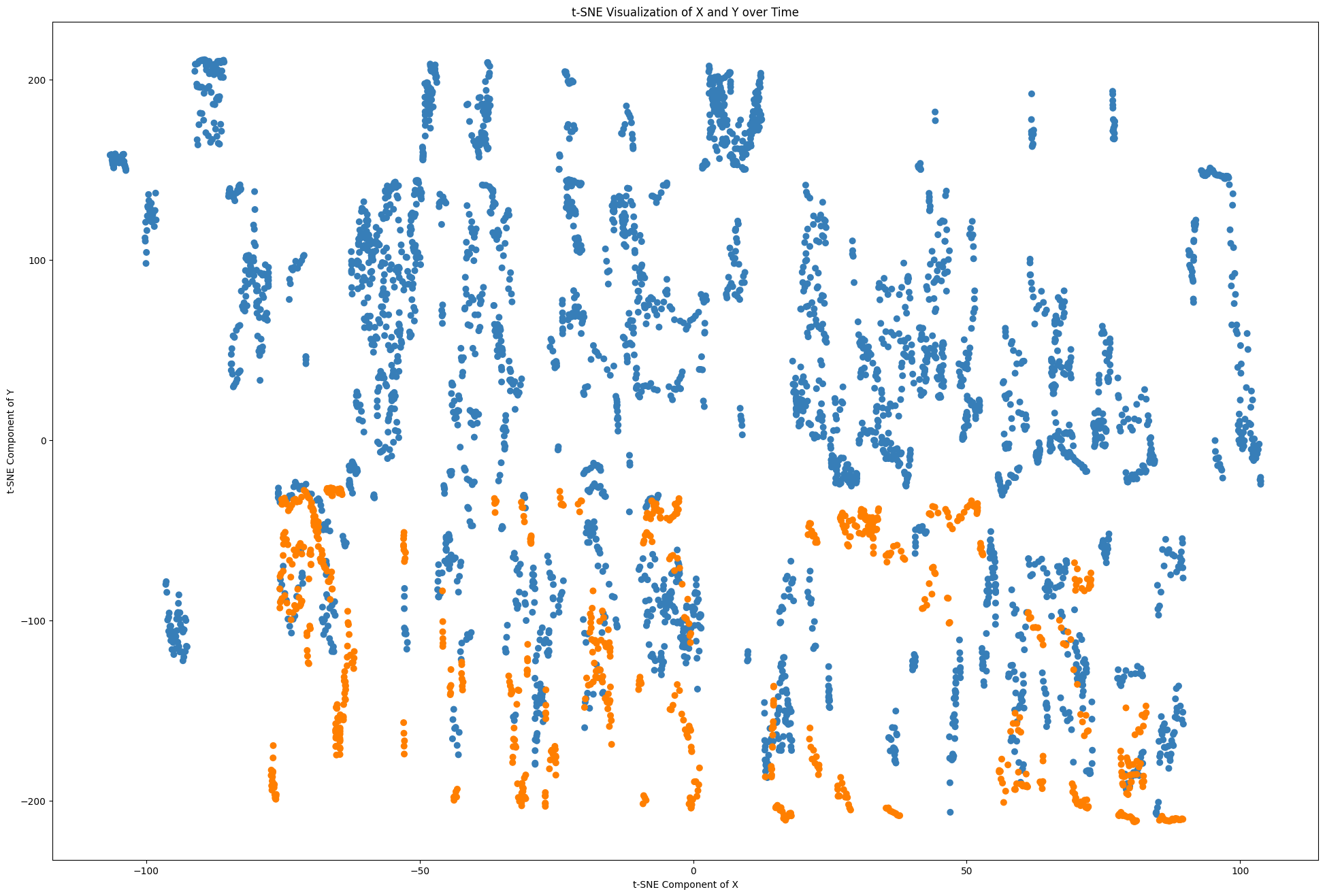}
        \captionsetup{font=small}
        \caption{FXB}
        \label{fig:sub2}
    \end{subfigure}%
    \hspace{0.01\textwidth}
    \vspace{-0.3cm}
    \captionsetup{font=small}
    \caption{State representation (x-axis) and reward (y-axis) reduced to one dimension via t-SNE (Algorithm \ref{alg:tsnegeneration} in Appendix G). The shift in distribution between training (blue points) and testing (orange points) highlights the out-of-distribution issue during testing.}
    \label{fig:T-SNE_comparision}
    \vspace{-0.5cm}
\end{figure}

Algorithmic trading systems have become an essential component of financial markets, with reinforcement learning (RL) emerging as a promising method for making financial decisions \cite{sun2023trademaster}. However, while these systems often excel in learning from large volumes of historical data, they usually fail to maintain similar performance in out-of-sample data. This overfitting arises from the highly dynamic nature of financial markets, where the testing dynamics diverge from the training dynamics, as illustrated in Fig \ref{fig:T-SNE_comparision}, because markets are continually reshaped by shifting macroeconomic indicators—such as interest rates and inflation \cite{mohammad2009impact}—and complex interactions among market participants.

% This overfitting stems in part from the highly dynamic nature of financial markets, the testing dynamics drift from the training dynamics as illustrated in Fig \ref{fig:T-SNE_comparision}, which are continually influenced by shifting macroeconomic indicators—such as interest rates and inflation \cite{mohammad2009impact} as well as complex interactions among market participants.

% \begin{figure}[htbp]
% \vspace{-0.2cm}
% \captionsetup{skip=2pt}
%   \centering
%     \includegraphics[width=0.30\textwidth]{figures/macro_ic.png}
%     \caption{Correlation matrix linking major macroeconomic indicators (rows) to various financial instruments (columns).
%     \red{simin: ugly and make no sense. inappropriate for intro image. use some img that captures the eye of reviewers in one second}
%     }
%     \label{fig:macro_ic}
% \vspace{-0.3cm}
% \end{figure}

% Most existing research in this domain focuses on equity markets and cryptocurrencies \cite{sun2023trademaster}, where backtests can appear promising but often downplay the magnitude of macroeconomic drivers. By contrast, our work centers on Exchange-Traded Funds (ETFs) of commodities, foreign exchange (FX) pairs, and stock indices—markets that are inherently tied to broader economic conditions.

% Despite macroeconomic drivers play a substantial role in shaping financial returns, 

% where backtests can appear promising but often understate these broader economic forces

Much of the existing research in this domain centers on equity markets and cryptocurrencies \cite{sun2023trademaster}. By contrast, our work focuses on Exchange-Traded Funds (ETFs) of commodities, foreign exchange (FX) pairs, and stock indices—important financial instruments whose performance is more directly correlated with global economic conditions \cite{tang2012index,engel2016exchange}. These instruments exhibit substantial correlations with macroeconomic indicators, providing a strong signal for trading and risk management. In finance research, authorities like the Federal Reserve emphasize macro-driven stress testing; for example, the Dodd-Frank Act Stress Tests \cite{FRB_DFAST_2024} require evaluating resilience under severe economic scenarios. While these frameworks are well-established in standard practice, they remain under-explored in algorithmic trading, highlighting the need for more robust, macro-sensitive approaches.

% One way to address the misalignment between backtesting and real-world trading is by adopting robust decision-making frameworks that explicitly incorporate uncertainty and adversarial market conditions. Traditional machine learning methods often overfit to historical distributions and assume that future conditions will mirror the past. By contrast, robust reinforcement learning (RL) \cite{nilim2005robustMDP3, iyengar2005robustMDP4} acknowledges that the true market environment can deviate significantly from historical patterns. Formulated under robust Markov Decision Processes (MDPs), robust RL employs a min-max paradigm in which trading strategies seek to maximize returns under a range of adversarial conditions. Yet the spectrum of potential “adverse” regimes can be vast, particularly when macroeconomic shifts that are unobserved in historical samples come into play. 

% \red{simin: are we doing RL from the very begining? or only introduce it here? are we not trading by RL?}

One way to address this is by adopting robust decision-making frameworks that explicitly incorporate uncertainty and adversarial market conditions. Machine learning methods often overfit to historical distributions and assume that future conditions will mirror the past. By contrast, robust reinforcement learning (RL) \cite{nilim2005robustMDP3, iyengar2005robustMDP4} acknowledges that real-world environment can deviate significantly from simulations, employing a min-max paradigm in which policy maximizes return under worst-case uncertainties. However, existing robust RL approaches are not tailored for trading and fail to account for unobservable macroeconomic drivers that cause drastic regime shifts, limiting their effectiveness. Specifically, implementing robust RL in financial markets under unobservable macroeconomic shifts entails two main challenges: I) developing a data generator that produces realistic, diverse market trajectories conditioned on macro indicators; and II) designing a robust RL framework capable of optimizing trading decisions under adversarial macroeconomic scenarios.

Generating realistic and diverse adversarial conditions is non-trivial. While generative models \cite{yoon2019time,ni2020conditional} can be employed to synthesize counterfactual data, it is hard to capture the complexity of financial market dynamics. We propose a macro-conditioned hybrid generator based on a Generative Adversarial Network (GAN) that captures temporal, inter-instrument, inter-feature, and feature–macro correlations in financial markets—key features that characterize the inherent structure of financial market data. By conditioning on macroeconomic indicators, our model learns from historical data while systematically injecting macro-driven variations. This design enables the synthesis of realistic market scenarios covering a broad range of potential “adverse” regimes that go beyond history. Unlike methods treating macro variables as ancillary features, we make them primary controllable conditions in data generation, yielding a more diverse training environment. 
% \red{add related work not addressing the problem}
% enriching the training of robust RL strategies with diverse, plausible environments.
% To generate diverse and realistic “what-if” market scenarios that stress-test such robust RL methods, deep generative models offer a compelling method \cite{yoon2019time,ni2020conditional}. 

% By conditioning synthetic market trajectories on inflation, interest rates, and other macro indicators, we can explore regime shifts that surpass those seen in historical datasets.

% Building upon these ideas, we introduce a \textbf{Bayesian Robust Framework} to systematically integrate macroeconomic uncertainties into the training and evaluation of trading strategies. Conceptually, we formulate the problem as a two-player zero-sum Bayesian game:

% \begin{itemize} \item \textbf{Adversarial Agent}: Controls a data generator parameterized by macroeconomic indicators. By perturbing these parameters away from historical norms, the adversary creates “worst-case” or stress-inducing market conditions. \item \textbf{Trading Agent}: Learns and updates trading policies based on observed market trajectories and its evolving beliefs about market states. The trading agent seeks robust profitability across a wide range of potential economic regimes. \end{itemize}

Building upon these, we introduce a Bayesian Robust Framework to systematically integrate macroeconomic uncertainties into the training of the trading policy. Conceptually, we formulate the problem as a two-player zero-sum Bayesian Markov game in which worst-case macroeconomic uncertainties are modeled as an adversarial agent that controls macroeconomic indicators in our data generator. By perturbing them away from historical norms, the adversary induces “worst-case” or stress-inducing market conditions. 
Meanwhile, the trading agent is formulated as a defender that maximize its profit under observed market conditions and Bayesian beliefs about the unknown macroeconomics in current market. By explicitly optimizing for worst-case robustness, the framework maximize the performance lower bound for the trading agent, ensuring its actual returns consistently exceed this threshold and remain reliable under highly volatile or adversarial conditions.

From a theoretical perspective, we aim to achieve a Robust Perfect Bayesian Equilibrium (RPBE) \cite{fudenberg1991perfect}, in which 
the trading agent’s policy is optimal given its belief over the market, and the adversary’s perturbations 
capture worst-case macroeconomic scenarios. To solve for this equilibrium, we adopt Bayesian neural fictitious 
self-play (Bayesian NFSP) \cite{heinrich2016deep}, enabling stable learning dynamics by computing max-min optimization with time-averaged opponent policy. Meanwhile, the Bayesian extension ensures 
that the trading agent maintains and updates a belief distribution over market states, enhancing its 
adaptability to adversarial macroeconomic shifts. 

% \red{add quantile belief newtwork}

Overall, our key contributions are: 
\begin{itemize} \item \textbf{Adversarial Framework for Macroeconomic Uncertainties.} We cast trading as a two-player zero-sum game, where a trading agent makes trading decisions and an adversarial agent perturbs macroeconomic conditions. \item \textbf{Macro-Conditioned Data Generation.} We propose a deep generative model that synthesizes market data conditioned on macroeconomic indicators. \item \textbf{Bayesian Game-Theoretic Optimization.} We cast the robust trading agent and worst-case agent as a two-player zero-sum Bayesian Markov game, solving for a Robust Perfect Bayesian Equilibrium via a quantile belief network and Bayesian neural fictitious self-play. \item \textbf{Enhanced Robustness and Profitability.} Experiments across 9 ETFs against 9 baselines show our method consistently gains higher profit with lower risk, and offer robust solution in extreme events like the COVID pandemic. \end{itemize}

\section{BACKGROUND AND RELATED WORKS}

\subsection{Robust RL}

Our research lies within the field of robust reinforcement learning (RL), which is theoretically grounded in the robust Markov Decision Process (MDP) framework \cite{nilim2005robustMDP3, iyengar2005robustMDP4, tamar2013robustMDP6, wiesemann2013robustMDP5}. In a robust MDP, the agent (defender) is trained to be resilient against an adversary that selects worst-case scenarios within an uncertainty set. These uncertainties can arise in environment transitions \cite{pinto2017rarl, mankowitz2019robustMDP1, xie2022robust}, actions \citep{tessler2019actionrobustmannor}, states \cite{zhang2020samdp, zhang2021atla}, or rewards \cite{wang2020robustreward}. Our work extends this framework by introducing dynamic, time-varying perturbations to the environment while enabling the trading agent to identify and adapt to these attacks for improved performance. Additionally, our research is related to adversarial policy learning \cite{gleave2019iclr2020advpolicy, guo2021icml2021}, which demonstrates that RL agents can be exploited by adversarial agents executing learned worst-case policies. In our setup, the adversary serves as a type of adversarial policy designed to maximally exploit the vulnerabilities of trading agents. However, unlike traditional adversarial policy research, our primary goal is to train a trading agent that achieves robustness without direct observation of the perturbations introduced by the adversarial policy.

% Our research lies within the field of robust reinforcement learning (RL), which is theoretically grounded in the robust Markov Decision Process (MDP) framework \cite{nilim2005robustMDP3, iyengar2005robustMDP4, tamar2013robustMDP6, wiesemann2013robustMDP5}. In a robust MDP, the agent (defender) is trained to be resilient against an adversary that selects worst-case scenarios within an uncertainty set. These uncertainties can arise in environment transitions \cite{pinto2017rarl, mankowitz2019robustMDP1, xie2022robust}, actions \citep{tessler2019actionrobustmannor}, states \cite{zhang2020samdp, zhang2021atla}, or rewards \cite{wang2020robustreward}. Our work extends this framework by introducing dynamic, time-varying perturbations to the environment while enabling the trading agent to identify and adapt to these attacks for improved performance. Additionally, our research is related to adversarial policy learning \cite{gleave2019iclr2020advpolicy, guo2021icml2021}, which demonstrates that RL agents can be exploited by adversarial agents executing learned worst-case policies. In our setup, the adversary serves as a type of adversarial policy designed to maximally exploit the vulnerabilities of trading agents. However, unlike traditional adversarial policy research, our primary goal is to train a trading agent that achieves robustness without direct observation of the perturbations introduced by the adversarial policy.

\subsection{Bayesian Game}

First introduced by Harsanyi, Bayesian games provide a theoretical framework for analyzing games with incomplete information \cite{harsanyi1967games}. In a Bayesian game, each agent is unaware of the types of other agents and must act optimally based on its beliefs about others, leading to the formation of a Perfect Bayesian Equilibrium. Agents update their beliefs about others' types using Bayes' rule, incorporating observations of others' actions. Applications of Bayesian games include ad hoc coordination in multi-agent reinforcement learning (MARL), where they facilitate the coordination of agents with varying preferences and types \cite{albrecht2015adhoc1, albrecht2016adhoc2, stone2010adhoc, barrett2017adhoc}. Bayesian games have also been applied to robust MARL \cite{xie2022robust, li2023byzantine}, particularly for identifying attackers with unknown types. However, these studies assume that each player's true type remains static. In our financial trading setting, the true economic states are both hidden and evolve dynamically over time. Trading agents must infer these evolving factors from their observations, introducing additional challenges.

% First introduced by Harsanyi, Bayesian games provide a theoretical framework for analyzing games with incomplete information \cite{harsanyi1967games}. In a Bayesian game, each agent is unaware of the types of other agents and must act optimally based on its beliefs about others, leading to the formation of a Perfect Bayesian Equilibrium. Agents update their beliefs about others' types using Bayes' rule, incorporating observations of others' actions. Applications of Bayesian games include ad hoc coordination in multi-agent reinforcement learning (MARL), where they facilitate the coordination of agents with varying preferences and types \cite{albrecht2015adhoc1, albrecht2016adhoc2, stone2010adhoc, barrett2017adhoc}. Bayesian games have also been applied to robust MARL \cite{xie2022robust, li2023byzantine}, particularly for identifying attackers with unknown types. However, these studies assume that each player's true type remains static. In our financial trading setting, the true economic states are both hidden and evolve dynamically over time. Trading agents must infer these evolving factors from their observations, introducing additional challenges.

\subsection{Financial Time-series Generation}
A challenge in algorithmic trading is to provide high-quality and diverse market data. Traditional agent-based methods simulate market participants to replicate the data stream \cite{axtell2022agent} or integrate stochastic effects \cite{shi2023neural}, are often constrained by assumptions about agent behaviors and empirical models, raising doubts about their capacity to capture real-world complexity \cite{vyetrenko2020get}. In contrast, deep generative models provide a data-driven alternative, learning temporal and cross-sectional dependencies from data. Generative Adversarial Networks (GANs) have emerged as a popular framework: TimeGAN \cite{yoon2019time} combines autoencoder-based representations with adversarial training. In the financial realm, FIN-GAN \cite{takahashi2019modeling} employs a vanilla GAN architecture to synthesize price features. Conditioning these generative models on macroeconomic factors offers an avenue to simulate market regimes and tail events not fully observed in historical datasets, thereby improving stress-testing and enhancing the robustness of trading. While DeepClair \cite{choi2024deepclair} integrates a FEDformer \cite{zhou2022fedformer} into RL to improve portfolio returns through better trend prediction, our method fundamentally differs by using a GAN-based generator for market simulation within a Bayesian macro-adversarial game. This design is motivated by the need for an adversarial generator, rather than a forecasting backbone.

% A challenge in algorithmic trading research is to provide high-quality and diverse market data for deep learning models. Traditional agent-based methods, which simulate individual market participants to replicate the data stream \cite{axtell2022agent} or integrate stochastic effects \cite{shi2023neural}, are often constrained by assumptions about agent behaviors and limited empirical models, raising doubts about their capacity to capture real-world complexity \cite{vyetrenko2020get}. In contrast, deep time-series generative models provide a data-driven alternative, learning intricate temporal and cross-sectional dependencies directly from historical records without imposing strong behavioral assumptions. Generative Adversarial Networks (GANs) have emerged as a popular framework in this domain: TimeGAN \cite{yoon2019time} combines autoencoder-based representations with adversarial training. In the financial realm, FIN-GAN \cite{takahashi2019modeling} employs a vanilla GAN architecture to synthesize price features—albeit in a more rudimentary form compared with recent time-series methods. Conditioning these generative models on macroeconomic factors offers an avenue to simulate elusive market regimes and tail events not fully observed in historical datasets, thereby improving stress-testing and enhancing the robustness of algorithmic trading strategies.

\section{PROBLEM FORMULATION}
We begin by introducing the Markov Decision Process (MDP) framework that defines our trading agent and its associated tasks. Subsequently, we outline the formulation of our market data generator and adversarial agent, which are designed to generate synthetic training data to enhance the robustness of the trading agent.

\begin{figure*}[htbp]
\vspace{-0.2cm}
\captionsetup{skip=2pt}
  \centering
    \includegraphics[width=0.8\textwidth]{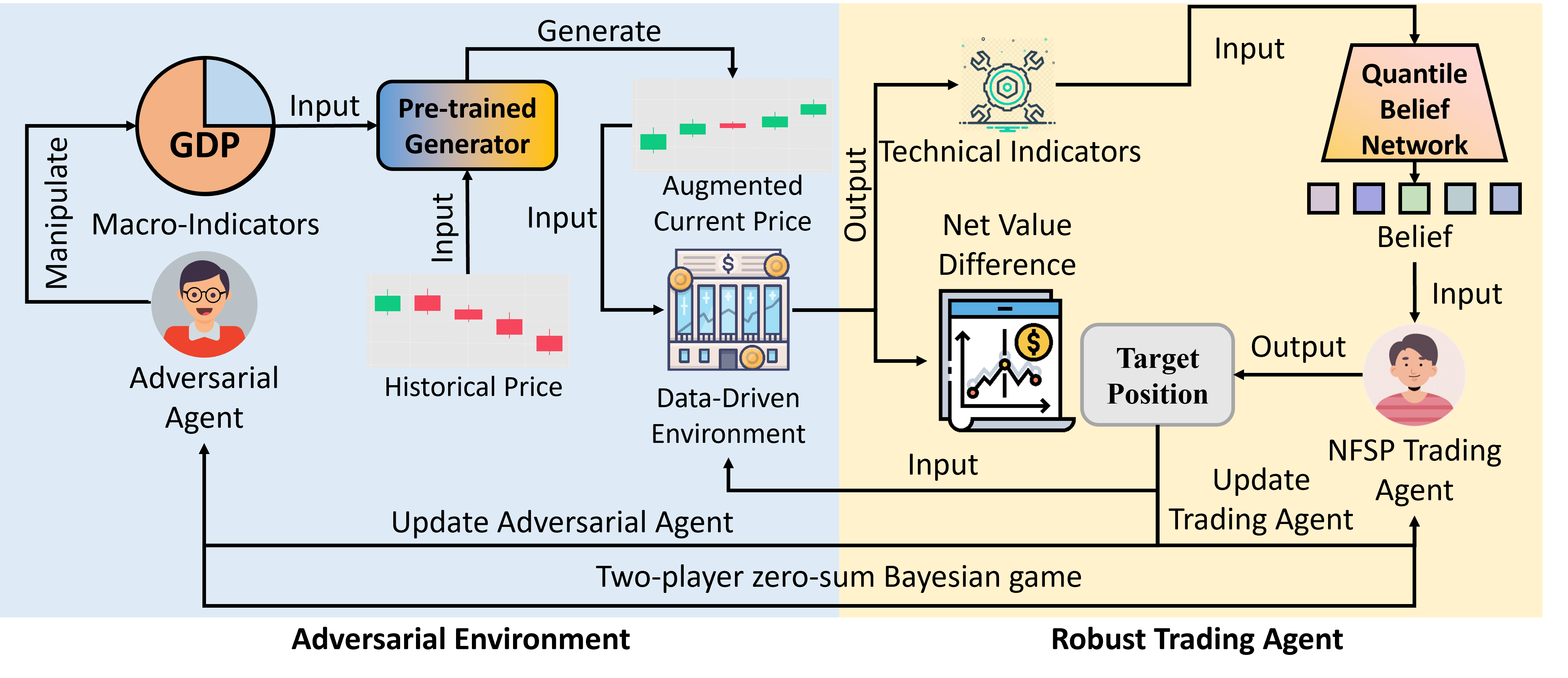}
    \caption{
    Overall architecture of our framework. An adversarial agent and a pre‐trained generator jointly form an adversarial environment, while a trading agent is trained on observations augmented by this environment.}
    \label{fig:framework}
\vspace{-0.3cm}
\end{figure*}
\subsection{Financial Market Formulation}
\label{sec:generator_formulation}

In this section, we present our formulation for generating synthetic market data. 
Consider a market dataset $\mathbf{X}_{t}$ of length $L$ at time $t$. This dataset comprises observations of multiple financial instruments and their associated features (e.g., price, volume), yielding a three-dimensional tensor of $(\text{timesteps}, \text{instrument}, \text{feature})$. Our aim is to learn a probabilistic model that can generate synthetic samples $\mathbf{X}_{t}'$ with realistic statistical properties, including: I) Temporal correlations; II) Inter-instrument correlations; III) Inter-feature correlations; IV) Feature-macro correlations.
% \begin{enumerate}
%     \item \textbf{Temporal Dependencies}: Financial time series typically exhibit autocorrelation structures (e.g., volatility clustering, seasonality, and momentum). Our model must capture such time-based dependencies across multiple scales.
%     \item \textbf{Inter-Instrument Correlations}: The returns, prices, or other characteristics of different financial instruments often move together due to shared market conditions, sector-wide events, or macroeconomic factors. Capturing these cross-instrument relationships is essential for building realistic market scenarios.
%     \item \textbf{Inter-Feature Relationships}: Within a single instrument, different features (e.g., bid price, ask price, volume) are related in ways that reflect underlying market microstructure. Preserving these relationships is important for applications such as liquidity modeling and order-book simulations.
% \end{enumerate}

% We define our generative model $G$ to produce synthetic market data conditioned on both macroeconomic states and past market observations, while incorporating stochastic noise. Specifically, let
We define our generative model $G$ as
\[
G\bigl(\mathbf{M}_{t,t-L}, \mathbf{N}, \mathbf{X}_{t-L}\bigr) \;=\; \mathbf{X}_{t}',
\]
where:
\begin{itemize}
    \item $\mathbf{X}_{t}'=[\mathbf{x}_{t-L+1}',...,\mathbf{x}_{t}']$ is the synthetic data of length $L$ at time $t$, the abbreviation for $\mathbf{X}_{[t-L+1:t]}'$.
    \item $\mathbf{M}_{t,t-L}==[\mathbf{m}_{t-2L+1},...,\mathbf{m}_{t}]$ is the macroeconomic state observed over the interval $[t-2L+1, t]$, which has length $2L$, the abbreviation for for $\mathbf{M}_{[t-L+1:t],[t-2L+1:t-L]}$. 
    \item $\mathbf{X}_{t-L}$ is the historical market data of length $L$ at time $t-L$, the abbreviation for $\mathbf{X}_{[t-2L+1:t-L]}=[\mathbf{x}_{t-2L+1},...,\mathbf{x}_{t-L}]$. 
    \item $n\in\mathbf{N}$ is a random noise variable.
\end{itemize}
% Formally, a single generated sample $\mathbf{x}_{t}'$ can be written as:
% \[
% \mathbf{x}_{t}' \;=\; G\bigl(\mathbf{m}_{t,t-L}, \mathbf{n}, \mathbf{x}_{t-L}\bigr),
% \]
% where $\mathbf{m}_{t,t-L}$ is the observed macro data over $[t-2L, t]$, $\mathbf{n}$ is drawn from $\mathbf{N}$, and $\mathbf{x}_{t-L}$ is the observed historical market data at time $t-L$.

\subsubsection{Distribution Approximation}
Our generative model $G$ is tasked with learning the conditional distribution
\[
\hat{p}\bigl(\mathbf{X}_{t} \;\big|\; \mathbf{M}_{t,t-L},\, \mathbf{X}_{t-L}\bigr),
\]
% aiming to approximate the true conditional distribution
% \[
% p\bigl(\mathbf{X}_{t} \;\big|\; \mathbf{M}_{t,t-L},\, \mathbf{X}_{t-L}\bigr).
% \]
From a joint distribution perspective, we can write:
\[
\begin{aligned}
p\bigl(\mathbf{X}_{t},\, \mathbf{M}_{t,t-L},\, \mathbf{N},\, \mathbf{X}_{t-L}\bigr)
&=\; p\bigl(\mathbf{M}_{t,t-L},\, \mathbf{X}_{t-L}\bigr)\,p\bigl(\mathbf{N}\bigr) \\
&\quad \times\; p\bigl(\mathbf{X}_{t} \;\big|\; \mathbf{M}_{t,t-L},\, \mathbf{N},\, \mathbf{X}_{t-L}\bigr),
\end{aligned}
\]
where $\mathbf{N}$ is assumed to be independent of the other variables. 
% The goal of the training procedure is to adjust $G$ so that samples drawn from $\hat{p}(\cdot)$ closely mirror the properties of samples from the true $p(\cdot)$. In practice, this involves optimizing a learning objective---such as an adversarial loss in a Generative Adversarial Network (GAN) framework or a maximum likelihood estimate in a Variational Autoencoder (VAE)---to align the distributions as closely as possible.

\subsubsection{Auto-Regressive Transaction Distribution}
Market data for a single time step $t$ can often be broken down into smaller ``ticks,'' indexed by $k \in \{1, 2, \ldots, K\}$. Each tick $\mathbf{X}_{t,k}$ may correspond to a transaction, a quote update, or another micro-event within the time interval $[t, t+1)$. To capture the finer granularity and the inherently auto-regressive nature of these events, we factorize the distribution of $\mathbf{X}_{t}$ at each tick $k$ conditionally on the previous ticks within the same time step. Formally,
\[
\begin{aligned}
p\bigl(\mathbf{X}_{t,1:K} \;\big|\; \mathbf{X}_{t-L},\, \mathbf{M}_{t,t-L},\, \mathbf{N}\bigr)
&= \prod_{k=1}^{K}
   p\Bigl(\mathbf{X}_{t,k} \,\Big|\,
     \mathbf{X}_{t,1},\ldots,\mathbf{X}_{t,k-1}, \\
&\qquad\quad \mathbf{X}_{t-L},\, \mathbf{M}_{t,t-L},\, \mathbf{N}\Bigr).
\end{aligned}
\]
% This explicit factorization captures intra-interval dependencies---such as the way one transaction's price or volume might influence the next transaction---while still conditioning on the historical market state $\mathbf{X}_{t-L}$ and macroeconomic variables $\mathbf{M}_{t,t-L}$. The random noise $\mathbf{N}$ again helps ensure diversity in the generated ticks.

% \subsection{Summary and Rationale}
% By unifying macroeconomic data, historical market observations, and noise, our proposed generative model is structured to capture:
% \begin{itemize}
%     \item \textbf{Long-Term Historical Patterns} through conditioning on $\mathbf{X}_{t-L}$.
%     \item \textbf{Broader Market Trends} and exogenous influences via $\mathbf{M}_{t,t-L}$.
%     \item \textbf{Intra-Interval Dynamics} through an auto-regressive formulation of sub-ticks or micro-events $\mathbf{X}_{t,k}$.
%     \item \textbf{Stochastic Variation} via the noise variable $\mathbf{N}$, thereby promoting diverse and plausible market scenarios.
% \end{itemize}

% This rigorous formulation lays the foundation for training procedures that align generated samples $\hat{p}$ with the true market distribution $p$. In practical applications---such as systematic trading, risk management, or synthetic data augmentation---this approach enables more robust and realistic simulations than naive or simplified time-series models, thus facilitating improved decision-making and scenario analysis.

\subsection{Robust Trading Agent}

\subsubsection{RL agents for trading.} With the definition of data generator, we can now give the formal definition of agents trading a single financial instrument, which is an Exchange-Traded Funds (ETFs) of commodities, foreign exchange pairs, and stock indices. Conventionally, The trading problem can be framed as a sequential decision-making process, where an agent seeks to maximize total profit under uncertainties. This problem is naturally modeled as a trading Markov Decision Process (MDP) within the reinforcement learning (RL) framework \cite{sun2023trademaster}. In this setup, the agent interacts with the market by making trading decisions. Formally, the MDP is defined by the tuple $\langle\mathcal{S}, \mathcal{A}, \mathbf{M}, \mathcal{T}, \mathcal{R}, \gamma\rangle$.
\begin{itemize}
    \item The \textbf{state space} $\mathcal{S}$ consists of a set of technical indicators and the agent's position. At time $t$, the state is defined as $s_t = [f(X_{[t-L+1,t]}), X_{[t]}, a_{t-1}]$, where $f(\cdot)$ is the operator that transforms a chunk of data into technical indicators and $a_{t-1}$ is the previous action taken by the agent.
    \item The \textbf{action space} $\mathcal{A}$ includes three possible choices: long position, short position, or close position \cite{deng2016deep}.
    \item The \textbf{macroeconomic indicator} \textbf{M} is a market representation, which is correlated with the state transition function.
    \item The \textbf{state transition function} $\mathcal{T}: \mathcal{S} \times \mathbf{M} \times \mathcal{A} \times \mathcal{S} \rightarrow [0, 1]$ describes how market states evolve over time. The transition function can be approximated by the learned data generator.
    \item The \textbf{reward function} $R: \mathcal S \times \mathcal A \rightarrow \mathbb R$ is computed as the return of the net value, incorporating factors such as transaction fees \cite{wang2021commission}. 
    \item $\gamma \in [0, 1)$ is the discount factor.
\end{itemize}
The trading process continues in $T$ steps. In each step $t \in T$, the agent executes an action $a_t$ using a policy $\pi(a_t|s_t)$, and observes reward $r_t$. The goal is learn a policy $\pi^*$ to maximize the value function $V_\pi(s_{0}) = \mathbb{E}_{\pi}[\sum_{t=0}^T \gamma^t r_t|s_t = s]$, which satisfies the following Bellman equation:
\[
V_\pi(s) = R(s, a) + \gamma \sum_{s' \in \mathbf{S}} \hat{p}\bigl(f(\mathbf{X}_{t}) \;\big|\; \mathbf{M}_{t,t-L},\, f(\mathbf{X}_{t-L})\bigr)V_\pi(s').
\]

\subsubsection{Threat Model.} In live trading, the macroeconomic data are updated at a slower frequency than the decision process of the trading agent. Such a slowly-evolving macroeconomic factor is a reflection of a longer process and does not reflect the accurate macroeconomic situation encountered by a trading agent that operates at a faster frequency. Since the ground truth macroeconomic data $\mathbf M^{*}_{t-L}$ is not known to the trading agent, we assume the slowly-evolving macroeconomic data $\mathbf M_{t, t-L}$ deviates from the ground truth data by $\mathbf M^{\alpha}_{t, t-L}$, satisfying the following conditions:
\[
\mathbf M^{*}_{t-L} = \mathbf M_{t, t-L} + \epsilon \mathbf M^{\alpha}_{t-L},
\]
where $\epsilon$ is a hyperparameter to measure the extent of changes in observed macroeconomic data. In our paper, we use an adversarial RL agent to learn the worst-case change in macroeconomic data that minimize the reward of the trading agent, defined as:
\[
\mathbf M^{\alpha, *}_{t-L} \in \argmin_{M^{\alpha}_{t-L}} V_\pi(s).
\]

\subsubsection{Solution concept.} Under the worst-case perturbations applied on the macroeconomic factors, our robust financial trading agent $\pi(\cdot|s_t)$ must be able to maximize profit under such uncertainties:
\[
\pi^*(\cdot|s_t) \in \argmax_{\pi} \min_{M^{\alpha}_{t-L}} V_\pi(s).
\]

% After 

% After the policy $\pi^*$ is learned, we deploy the policy in real market with unknown dynamics $\mathcal{T}$ to test its performance.

% We test the performance of $\pi^*$ in markets whose dynamics $\mathcal{T}$ are unknown.

% Note that the ``Perfect'' concept of equilibrium requires the trading agent to perform optimally according to the information gathered at the current timestep, and do not it to perform optimally at hindsight.
% Note that the “Perfect” concept of equilibrium requires the trading agent to act optimally based on the information available at the current timestep, rather than assuming optimal decisions with the benefit of hindsight.

\section{METHOD}

While the trading agent seeks to maximize reward despite market fluctuations, the ground-truth macroeconomic factors are rapidly changing and unobservable. As a result, there is an inherent level of incomplete information that conventional robust RL methods must incorporate to make effective trading decisions. This challenge remains unresolved in many existing approaches, which often assume stationary or fully observable market conditions.

In response to these challenges, we propose an adversarial framework, illustrated in Figure~\ref{fig:framework}, that comprises a pre-trained data generator, an adversarial agent, and a robust trading agent with a quantile belief network. We begin by describing our data generator for synthetic market data generation. Subsequently, we outline the adversarial agent that perturbs the data distribution, followed by the robust trading agent designed to operate effectively under adversarial market conditions.

% Moreover, while the trading agent seeks to maximize reward despite market fluctuations, the ground-truth macroeconomic factors are rapidly changing and unobservable. Thus, conventional robust RL methods should additionally incorporate this incomplete information for optimal decision-making.

% In this section, we present our adversarial framework, as illustrated in Figure \ref{fig:framework}, which comprises a pre-trained data generator, an adversarial agent, and a trading agent \red{which additionally incorporate the changing and unobservable macroeconomic factors for optimal decision-making}
% . We begin by introducing our data generator for synthetic market data generation. We then describe the adversarial agent and the robust trading agent.

% To train robust trading policies, we formulate worst-case uncertainty in macroeconomics and the robust trading agent as a two-agent zero-sum Bayesian Markov game, where uncertainties in macroeconomic factors are unobservable to traders. The robust trading policy seeks a Robust Perfect Bayesian Equilibrium, which is achieved by our quantile belief network and Bayesian neural fictitious self-play.

\subsection{Data Generator}
To address the absence of a realistic and diverse environment for trading, we propose our generator, following the hybrid architecture in TimeGAN \cite{yoon2019time}, tailored to model temporal, inter-instrument, inter-feature, and feature-macro correlations.

% \begin{algorithm}[htbp]
% \caption{Correlation-Weighted Imputation}
% \label{alg:correlation_weighted_imputation}

% \textbf{Input:} Ticker set \( T = \{t_1, \ldots, t_N\} \), feature datasets \( \{D_t\}_{t \in T} \), correlation matrices \( \{C^{(f)}\}_{f \in \mathcal{F}} \).\\
% \textbf{Output:} Imputed datasets \( \{D_t\}_{t \in T} \).

% For each feature \( f \in \mathcal{F} \), process all tickers \( t \in T \). For a given ticker \( t \), every time index \( i \) where \( D_t(i, f) \) is missing\;
% Identify the set of valid tickers:
% \[
% V = \{ t' \in T \setminus \{t\} : D_{t'}(i, f) \text{ is available} \}
% \]
% \If{\( V \neq \emptyset \)}{
%     Compute weights \( w_{t'} = \exp(C^{(f)}(t, t')) \) for all \( t' \in V\)\;
%     Normalize weights:
%     \[
%     \tilde{w}_{t'} = \frac{w_{t'}}{\sum_{s \in V} w_s}
%     \]
%     Impute missing value:
%     \[
%     D_t(i, f) = \sum_{t' \in V} \tilde{w}_{t'} \cdot D_{t'}(i, f)
%     \]
% }

% \Return Updated datasets \( \{D_t\}_{t \in T} \)\;

% \end{algorithm}

\subsubsection{Data Transformation}
Instead of directly generating raw price and volume data, we transform the raw into a feature set comprising open-to-close return, close-to-close return, low-to-close ratio, high-to-close ratio, and $\log(\text{close} \times \text{volume})$, allowing the generator to learn a more structured distribution. Additionally, to handle missing data in financial instruments, we implement a market-aware imputation technique that leverages pairwise correlations, detailed in the Algorithm \ref{alg:correlation_weighted_imputation} in the Appendix \ref{section:Algorithms}. For each missing entry, we identify all tickers with valid data at that point, 
compute correlation-based weights, and use their weighted average values to fill in the gap. This preserves both the temporal dynamics and market structure, unlike conventional imputation methods like mean-filling.

\begin{figure}[htbp]
\vspace{-0.2cm}
\captionsetup{skip=2pt}
  \centering
    \includegraphics[width=0.45\textwidth]{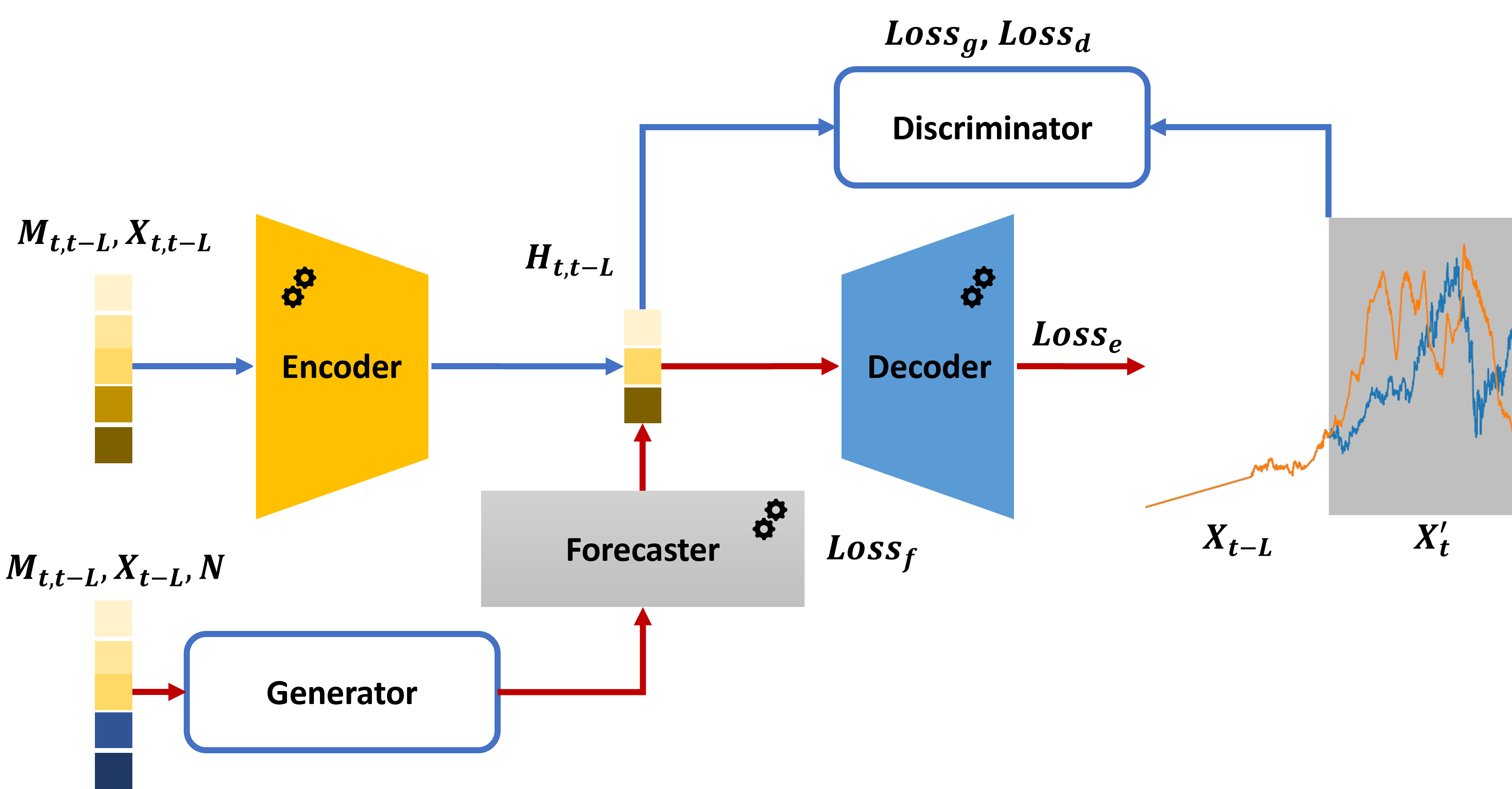}
    \caption{
    Overall architecture of the data generator. The encoder, decoder, and forecaster are pre-trained (marked with gear icons). The blue routes are only for training, while the red routes are for both training and inference.}
    \label{fig:generator}
\vspace{-0.3cm}
\end{figure}

% \subsubsection{Architecture}
% Following the hybrid generative adversarial network in TimeGAN \cite{yoon2019time} we proposed our generator to model temporal, inter-instrument, and inter-feature correlations as discussed in Section \ref{sec:generator_formulation}. An overall architecture of the data generator is shown in Figure \ref{fig:generator}.

% \textbf{AutoEncoder}
% % address how auto encoder help models inter-stock correlation and inter-feature correlation

% %introduce model design and loss

% \textbf{Forecaster}
% % address how forecaster helps model time correlation

% %introduce model design and loss

% \textbf{Generator and Discriminator}
% % address how loss helps generator captures mutiple key metrics of financial data

% %introduce model design and loss

\subsubsection{Architecture}
As shown in Figure~\ref{fig:generator}, the architecture of the proposed data generator comprises three key components: the encoder, forecaster, and decoder. These components are initially pre-trained on relevant tasks to capture the underlying features of the input time-series data. Subsequently, fine-tuning is performed with the inclusion of the generator and discriminator modules, enabling adversarial training for enhanced data generation. The detailed network architecture is in section \ref{section:network}.

\noindent
\textbf{AutoEncoder.}
To capture inter-instrument, inter-feature, and macro-feature correlations, we begin by passing the processed feature through an autoencoder. This module decomposes into an encoder that transforms high-dimensional features of multiple instruments and macro indicators into a compressed latent representation and a decoder that reconstructs features from this latent space. 
% By training the autoencoder on multiple instruments and multiple features simultaneously, the latent embeddings learned by the encoder preserve relationships such as co-movement among instruments as well as structural connections among features for a single instrument.

Formally, if $\mathbf{X}_t$ is the real market data at time $t$, then the encoder $E(\cdot)$ produces a latent embedding $\mathbf{H}_{t,t-L} = E(\mathbf{X}_{t,t-L},\mathbf{M}_{t,t-L})$. From there, a decoder $D(\cdot)$ attempts to reconstruct $\mathbf{X}_{t,t-L} \approx \tilde{\mathbf{X}}_{t,t-L} = D(\mathbf{H}_{t,t-L})$. The autoencoder training objective commonly adopts a mean squared error (MSE) loss:
\[
Loss_{\mathrm{e}} \;=\; \bigl\|\mathbf{X}_{t,t-L} - D\bigl(E(\mathbf{X}_{t,t-L},\mathbf{M}_{t,t-L})\bigr)\bigr\|^2,
\]
where $\|\cdot\|$ denotes an appropriate norm (e.g., $\ell^2$), possibly computed over all instruments and all features. Minimizing ${Loss}_{e}$ forces the latent representation $\mathbf{H}_{t,t-L}$ to preserve the cross-sectional and cross-feature patterns crucial for realistic data generation.

\noindent
\textbf{Forecaster.}
Financial time-series data exhibit certain temporal dynamics, including autocorrelation structures (e.g., volatility clustering and seasonality). To address this, we incorporate a forecaster that learns to predict the subsequent latent state given a sequence of past embeddings, taking the encoder’s latent outputs and generating a one-step-ahead prediction $\hat{\mathbf{H}}_{t+1}$. 
% By matching this prediction with the true latent embedding $\mathbf{H}_{t+1}$, the forecaster encourages the overall system to embed relevant temporal dependencies in $\mathbf{H}_t$. 

% Concretely, if $\mathbf{H}_t = E(\mathbf{X}_t)$ is the latent representation of the data at time $t$, the forecaster $F(\cdot)$ aims to satisfy
% \[
% \mathbf{H}_{t+1} \;\approx\; \hat{\mathbf{H}}_{t+1} \;=\; F\bigl(\mathbf{H}_t\bigr).
% \]
The corresponding loss is an MSE between forecasted and true latent representation, which is $
Loss_{\mathrm{f}} \;=\; \bigl\|\mathbf{H}_{t+1} - F(\mathbf{H}_t)\bigr\|^2.
$
By enforcing accurate forecasts in the latent space, the model learns to propagate temporal correlations through the generator.

\noindent
\textbf{Generator and Discriminator.}
% We adopt an adversarial mechanism inspired by GANs to produce realistic synthetic market trajectories. 
The generator $g(\cdot)$ outputs synthetic latent representations. Meanwhile, the discriminator $d(\cdot)$ differentiates real latent embeddings from those generated by $g$. 
% This section describes the main components and loss functions of both modules, illustrating how we align the synthetic data with the statistical properties of real financial time series.

\paragraph{Generator.}
The generator is trained to (1) fool the discriminator by making its synthetic outputs resemble real embeddings and (2) preserve essential market statistics:
\[
g\bigl(\mathbf{M}_{t,t-L}, \mathbf{N}, \mathbf{X}_{t-L}\bigr) \;=\; \mathbf{H}_{t,t-L}',
\]
% Specifically, it takes in noise tensors $\mathbf{N}$ and produces latent representations $$.

The forecaster then refines these latent outputs to capture temporal correlations. Finally, the decoder maps $\mathbf{H}_{t,t-L}'$ back into $\mathbf{X}_{t,t-L}'$ and $\mathbf{X}_{t}'$ is used to get $s'$ : 
In practice, the generator’s loss $Loss_g$ comprises several terms:

\begin{itemize}
    \item \textit{Adversarial Terms.} Let $d(\cdot)$ output the logit that indicates “real” vs.\ “fake”. The generator seeks to maximize $d$’s confusion by minimizing a binary cross-entropy (BCE) loss:
    \[
      Loss_\mathrm{adv} \;=\; 
      \mathrm{BCE}\bigl(d(\mathbf{H}', M), \mathbf{1}\bigr)
      \;+\;
      \mathrm{BCE}\bigl(d(F(\mathbf{H}'),M), \mathbf{1}\bigr),
    \]

    % \item \textit{Supervision Term.} The generator also receives feedback from a forecaster (or “supervisor”) module. If $\mathbf{H} = E(\mathbf{X}, T)$ denotes the latent embedding of real data, the supervisor generates a one-step-ahead prediction $\hat{\mathbf{H}}_{\mathrm{sup}}$ from $\mathbf{H}$. By matching $\hat{\mathbf{H}}_{\mathrm{sup}}$ to the subsequent real latent $\mathbf{H}_{t+1}$, the network learns to respect temporal structure:
    % \[
    %   \mathcal{L}_\mathrm{sup} \;=\; \bigl\|\hat{\mathbf{H}}_{\mathrm{sup}}[:,:-1,:] - \mathbf{H}[:,1:,:]\bigr\|^2.
    % \]
    % Penalizing this discrepancy forces the internal representations to encode realistic time dependencies.

    \item \textit{Moment-Matching \& Masking.} Since financial data contain missing entries and often exhibit distinct statistical signatures,
    % we impose additional losses to align the first two moments (mean and variance) of $\mathbf{X}_{\hat{}}$ with those of the real data $\mathbf{X}$. 
    we compute a masked mean and variance loss within each feature, ignoring missing values:
    % \[\mu_\mathrm{real},\;\sigma_\mathrm{real}^2,\quad
    %   \mu_\mathrm{fake},\;\sigma_\mathrm{fake}^2.
    % \]
    % The moment-matching penalty adds
    \[
      Loss_\mathrm{moments} 
      \;=\; \bigl|\mu_\mathrm{fake} - \mu_\mathrm{real}\bigr|
        \;+\; \Bigl|\sqrt{\sigma_\mathrm{fake}^2} - \sqrt{\sigma_\mathrm{real}^2}\Bigr|,
    \]
    while a separate term $Loss_\mathrm{std}$ measures relative variance differences. Both are applied only over valid (non-missing) positions, as indicated by a mask $\mathbf{M}$.

    \item \textit{Divergence and Mode-Seeking.} Additional divergence-based losses can further align synthetic and real distributions, while mode-seeking objectives encourage diversity in generated sequences. By comparing generator outputs from different noise samples $(\mathbf{N}_1,\mathbf{N}_2)$, the mode-seeking loss ensures that similar inputs do not always map to the same synthetic trajectory, fostering richer diversity.
\end{itemize}
By summing these terms with suitable weights, $Loss_{g}$
% \[
% \mathcal{L}_G \;=\; 
% \mathcal{L}_\mathrm{adv}
% \;+\; \lambda_\mathrm{sup}\,\sqrt{\mathcal{L}_\mathrm{sup}} 
% \;+\; \lambda_\mathrm{mom}\,\mathcal{L}_\mathrm{moments}
% \;+\; \dots,
% \]
balances the requirements of adversarial realism, temporal coherence, and matching of key financial statistics.
\paragraph{Discriminator.}

The discriminator \( d(\cdot) \) is designed to differentiate real latent representations from synthetic ones. The discriminator is trained using binary cross-entropy (BCE) loss:
\[
Loss_d \;=\; 
\mathrm{BCE}\bigl(d(\mathbf{H}),\,1\bigr)
\;+\;
\mathrm{BCE}\bigl(d(\mathbf{H^{'}}),\,0\bigr)
\;+\;\mathrm{BCE}\bigl(d(f(\mathbf{H^{'}}),\,0\bigr),
\]

\noindent
In conjunction, these modules collectively allow us to generate synthetic market scenarios that realistically reflect the multifaceted dependencies present in real-world financial data.

% \subsection{Bayesian Robust Trading Agent via Adversial Training}

% We use DQN 

% based on DQN we add

% \subsubsection{Quantile Belief Network.}

% Why bayesian works?
% % 不完备信息博弈
% partily observerd
% % 把整个市场当作对手，用ma5建模对手策略导致的distribution shift

% % use ma5 loss curve 
% \subsubsection{Learning robust meta-policy with diverse adversial environments with Neural Fictitious Self-play.}

% \subsubsection{Advrsarial Agent.}
% % 学习如何达到最差情况
% % 解minmax

% % 从思民那篇文章谈起，说明一下和如何配合belief

% #C64a11
\definecolor{FBest}{RGB}{198,70,17} % first best results
% #E4A304
\definecolor{SBest}{RGB}{228,163,4} % second best results
% #548235
\definecolor{TBest}{RGB}{84,130,53} % third best results

\begin{table*}[hbpt]
\renewcommand{\arraystretch}{1.2}
\footnotesize
\vspace{-0.2cm}
\setlength{\abovecaptionskip}{0cm}
\centering
\begin{threeparttable}
% \setlength{\tabcolsep}{4pt}
% \resizebox{\linewidth}{!}{
\begin{tabular}{lcccccccccccc}
\toprule
\multirow{2}{*}{Models} & \multicolumn{3}{c}{DBB} & & \multicolumn{3}{c}{GLD} & & \multicolumn{3}{c}{UNG} \\ 
\cmidrule{2-4} \cmidrule{6-8} \cmidrule{10-12}
& ARR\%$\uparrow$ & SR$\uparrow$ & MDD\%$\downarrow$ & & ARR\%$\uparrow$ & SR$\uparrow$ & MDD\%$\downarrow$ & & ARR\%$\uparrow$ & SR$\uparrow$ & MDD\%$\downarrow$ \\ 
\midrule
(1) Buy and Hold & 0.02 & 0.19  & 35.11& & 4.77 & 0.66 & 21.03& & -18.71 &-0.18 & 86.62  \\
\hline
(2) DQN & 13.59 & 1.08 & 33.33& & 5.35 & 0.82 & 12.81 & &18.55 & 0.93 & 53.86  \\
(3) Robust Trading Agent & 13.93 & \textcolor{SBest}{1.61} & 12.70& & \textcolor{FBest}{9.25} & \textcolor{SBest}{1.20} & 11.14 &  &25.46 & 1.05 & 63.87  \\
(4) Naïve Adversarial & 2.79 & 0.45 & 15.61& & 7.66 & 1.08 & 12.96 &  &  23.84  & 1.03 & 52.53 \\
(5) RoM-Q & \textcolor{SBest}{16.26} & 1.22 & 21.38& & 8.06 & 1.13 & 19.30 &  & \textcolor{SBest}{27.17} & \textcolor{SBest}{1.09} &  54.92  \\
(6) RARL & 6.53 & 0.83 & \textcolor{SBest}{11.87}& & 1.37 & 0.91 & \textcolor{FBest}{2.84} &  & 26.74 & 1.07 &  51.20  \\
(7) DeepScalper & 5.89 & 0.65 & 20.60 & & 6.37 & 1.03 & 11.86 & & 18.62 & 0.92 &  79.52  \\
(8) EarnHFT & 5.94 & 0.60 & 29.95 & & 7.23 &1.00 & 12.06 &  &  24.33  & 1.03 & 70.64 \\
(9) CDQN-rp & 4.68 & 0.60 & 30.29 & & 6.48 & 0.86 & 21.56 &  & 18.71   & 0.92 & \textcolor{FBest}{42.91}\\
\hline
(10) w/o adv agent  & 13.04 & 1.02 & 24.21 & & 7.75 & 1.09 & 15.53 & & 24.92 & 1.05 &  50.54  \\
(11) IPG & 2.55 & 0.37 & 25.33 & & 5.81 & 1.01 & 10.10 & & 16.79 & 0.88 &  62.99  \\
\hline
Ours & \textcolor{FBest}{26.03} & \textcolor{FBest}{1.86} & \textcolor{FBest}{11.00}& &\textcolor{SBest}{8.96} & \textcolor{FBest}{1.20} & \textcolor{SBest}{8.33} &  & \textcolor{FBest}{29.12} & \textcolor{FBest}{1.10} & \textcolor{SBest}{49.37}   \\
\midrule
Models & \multicolumn{3}{c}{SPY} & & \multicolumn{3}{c}{QQQ} & & \multicolumn{3}{c}{IWM} \\ 
% \cmidrule{2-4} \cmidrule{6-8} \cmidrule{10-12}
% & ARR\%$\uparrow$ & SR$\uparrow$ & MDD\%$\downarrow$ & & ARR\%$\uparrow$ & SR$\uparrow$ & MDD\%$\downarrow$ & & ARR\%$\uparrow$ & SR$\uparrow$ & MDD\%$\downarrow$ \\ 
\midrule
(1) Buy and Hold & 7.72 & 0.82 & 25.36 & & 7.92 &  0.71 & 35.62 &  & -3.28 & -0.05 & 33.13 \\
\hline
(2) DQN & 7.94 & 0.84 & 19.64 & & 7.48 & 0.68 & 35.54 & & 10.71 & 1.13 & \textcolor{FBest}{19.60} \\
(3) Robust Trading Agent & \textcolor{SBest}{9.62} & \textcolor{FBest}{1.29} & \textcolor{SBest}{13.34} & & 8.37 & 0.74  & 34.04 & & 12.96 & 1.03 & 23.90 \\
(4) Naïve Adversarial & 6.40 & 0.74 & 22.56 & & 9.14 & 0.78 & 36.16 & & 11.30 & 0.99 & \textcolor{SBest}{22.58} \\
(5) RoM-Q & 7.52 & 0.81 & 26.14 & & 9.02 & 0.77 & 33.96 & & 13.31 & 1.01 & 29.71 \\
(6) RARL & 8.05 & 0.89 & 16.13 & & 8.53 & 0.83 & 30.68 & & 8.16 & 0.76 & 25.71 \\
(7) DeepScalper & 5.42 & 0.76 & 16.54 & & 7.70 & 0.69 & 35.61 &  & 9.06  &  0.80   & 26.37 \\
(8) EarnHFT & 5.53 & 0.73 & 13.72 & &8.17 &0.72 & 35.62 &  & 11.25   &\textcolor{SBest}{1.22}  & 28.42 \\
(9) CDQN-rp & 4.74  & 0.57 & 25.36 & & 6.96 & 0.65 & 37.05 &  &  6.32  & 0.66 & 25.68 \\
\hline
(10) w/o adv agent & 7.23 & 0.78 & 17.63 & & \textcolor{SBest}{11.47} & 0.91 & \textcolor{SBest}{28.54} & & \textcolor{SBest}{15.32} & 1.20 & 27.55  \\
(11) IPG  & 5.63 & 0.69 & 24.32 & & 11.24 & \textcolor{SBest}{0.93} & 29.72 & & 13.43 & 1.11 & 31.58   \\
\hline
Ours & \textcolor{FBest}{12.31} & \textcolor{SBest}{1.20} & \textcolor{FBest}{12.03} & & \textcolor{FBest}{19.03}  & \textcolor{FBest}{1.29} & \textcolor{FBest}{24.71} & & \textcolor{FBest}{17.32} & \textcolor{FBest}{1.41} & 24.92 \\
\midrule
Models & \multicolumn{3}{c}{DBC} & & \multicolumn{3}{c}{FXY} & & \multicolumn{3}{c}{FXB} \\ 
% \cmidrule{2-4} \cmidrule{6-8} \cmidrule{10-12}
% & ARR\%$\uparrow$ & SR$\uparrow$ & MDD\%$\downarrow$ & & ARR\%$\uparrow$ & SR$\uparrow$ & MDD\%$\downarrow$ & & ARR\%$\uparrow$ & SR$\uparrow$ & MDD\%$\downarrow$ \\ 
\midrule
(1) Buy and Hold & 11.78 & 0.99 & 27.78 & & -9.18 & -1.73 & 31.79 & & -3.08 & -0.49 & 24.97 \\
\hline
(2) DQN & 11.78 & 1.00 & 27.68 & & 8.22 & 1.39 & 13.94 & & 3.30 & 0.87 & 12.75 \\
(3) Robust Trading Agent & 9.27 & 0.98 & 16.78 & & 1.63 & 0.38 & 22.32 & & 3.21 & 0.67 & 10.42 \\
(4) Naïve Adversarial & \textcolor{FBest}{15.70} & 1.28 & 19.95 & & 7.54 & 1.29 & 11.53 & & \textcolor{SBest}{5.00} & 0.94 & 11.77 \\
(5) RoM-Q & 13.00 & 1.06 & \textcolor{SBest}{16.76} & & \textcolor{SBest}{13.78} & 2.13 & 10.64 & & 2.82 & 0.61 & 9.23 \\
(6) RARL & 12.37 & 1.11 & 29.23 & & 8.04 & 1.56 & 9.65 & & 0.11 & 0.08 & \textcolor{SBest}{7.95} \\
(7) DeepScalper & 10.21  & 0.93 & 18.58 & & 9.67 & 1.85 & \textcolor{FBest}{5.74}  &  &  1.54  & 0.46 & 10.21 \\
(8) EarnHFT & 8.25 & 0.89 & 27.74 & & 11.77 & 1.81 & 11.25  &  &  3.65   &  0.76 & 10.76 \\
(9) CDQN-rp & 8.53 & 0.83 & 28.79 & & 5.99 & 1.12 & 10.59 &  &  2.70   & 0.62 & 16.92\\
\hline
(10) w/o adv agent & 12.77 & \textcolor{SBest}{1.31} & 21.83 & & 13.15	 & \textcolor{SBest}{2.43} & 6.55 & & 2.70 & 0.62 &  16.92  \\
(11) IPG  & 10.62 & 0.95 & 19.58 & & 11.39 & 1.83 & 13.47 & & 4.15 & \textcolor{SBest}{1.02} & 9.52  \\
\hline
Ours & \textcolor{SBest}{15.29} & \textcolor{FBest}{1.34} & \textcolor{FBest}{16.50} & &  \textcolor{FBest}{15.14} & \textcolor{FBest}{2.51} & \textcolor{SBest}{5.80} &  & \textcolor{FBest}{5.13} & \textcolor{FBest}{1.07} & \textcolor{FBest}{7.62}\\
\bottomrule
\end{tabular}
% }
\end{threeparttable}
\caption{Performance Comparison of Trading Models: The best model is in \textcolor{FBest}{red}, the second-best in \textcolor{SBest}{yellow}.}
\label{tab:updated_results}
\vspace{-0.2cm}
\end{table*}

\subsection{Bayesian Robust Trading Agent via Adversarial Training}

Given the generator described above, we now focus on developing robust trading agent under adversarial market conditions. The goal is to maximize profit despite the unknown macroeconomic uncertainties. We formalize this as a Bayesian game, and define optimal solution as a robust perfect Bayesian equilibrium, which agents make optimal decisions based on current belief about the macroeconomic status of the market. To compute belief in highly complex market, we introduce a quantile belief network to simplify the inference of beliefs in practical trading settings. To achieve robust perfect Bayesian equilibrium, we propose Bayesian neural fictitious self-play, which provides a stable policy optimization framework that involves a robust trading agent and a worst-case adversary agent that controls the macroeconomic indicators.

\subsubsection{Bayesian Game Formulation.} Financial markets inherently provides incomplete information. The observable macroeconomic factor such as prices and volumes evolves at lower frequencies, and do not reflect the true macroeconomic factors at the time agents is making it decisions. Besides, many influential factors such as order flow, policy decisions remain partially hidden. Thus, while our policies are trained to be robust against worst-case uncertainties in macroeconomic factors, the policy do not have knowledge of the true macroeconomic factors at current time.

Bayesian game \cite{harsanyi1967games} provides a principled way of decision making under incomplete information. In a Bayesian game, the agent makes decisions based on beliefs about the incomplete information, which is updated by Bayes' rule as the game proceeds. Under the formulation of a Bayesian game, our trading agent does not treat the observed macroeconomic data $\mathbf M_{t, t-L}$ as ground truth and maintains a belief $b$ over the ground truth macroeconomic data through Bayes' rule, which infers the posterior distribution over macroeconomic indicators based on current market observations. 

% While such changes in worst-case uncertainties added to macroeconomic indicators are unknown to the trading agent, in our paper, we assume the trading agent does not treat the observed macroeconomic data $\mathbf M_{t, t-L}$ as ground truth and maintains a belief $b$ over the ground truth macroeconomic data through Bayes rule, which infers the posterior over macroeconomic factor based on current market observations. 
% \[
% b(\hat{\mathbf M}_{t, t-L}|x_{t}, x_{t-L}) = \frac{\hat{p}(x_t|\hat{\mathbf M}_{t,t-L},  \mathbf{X}_{t-L}) p(\hat{\mathbf M}_{t,t-L}| \mathbf{X}_{t-L}) }{\int_{M} \hat{p}(x_t|\mathbf M, \mathbf{X}_{t-L}) p(\mathbf M| \mathbf{X}_{t-L})  d\mathbf M}
% \]

Under the Bayesian game formulation, we treat the decision process as a \emph{two-agent zero-sum Bayesian Markov game}, where one adversary agent selects the worst-case uncertainties on macroeconomic data, simulating the worst case the trading agent can face. The robust trading agent must maximize its profit under such uncertainties, but have to infer the true perturbations added by the adversary. As such, the goal of the adversary is to learn a worst-case uncertainty added on macroeconomic factors, which minimize the profit gained by trading agent. In practice, we use an RL agent $\pi^\alpha(\mathbf M^{\alpha, *}_{t-L}|\mathbf M_{t-L}, S_{t-L})$ to learn such a worst-case agent, conditioned on the current state. Formally, this is defined as:
\[
\pi^\alpha(\mathbf M^{\alpha, *}_{t-L}| S_{t-L}) \in \argmin_{\pi^\alpha} V_\pi(s_t, b_t).
\]
where \(V_\pi(s_t, b_t)\) is the value function of the trading agent. Since in our Bayesian game formulation, the trading agent maintains a belief over the market, the adversary is required to manipulate the market subtly, such that the trading agent does not recognize an explicit change in the market but still performs suboptimally in trading. While the worst-case adversary can be optimized via any RL method, we train it via Q-learning.

% formulated as:
% Note that since the trading agent maintains a belief over the market, the adversary is required to manipulate the market subtly, such that the trading agent does not recognize an explicit change in the market but still performs suboptimally in trading. In practice, we use an RL agent $\pi^\alpha(\mathbf M^{\alpha, *}_{t-L}|\mathbf M_{t-L}, S_{t-L})$ to learn such a worst-case agent, conditioned on the current state. Formally, this is defined as:
% \[
% \pi^\alpha(\mathbf M^{\alpha, *}_{t-L}| S_{t-L}) \in \argmin_{\pi^\alpha} V_\pi(s_t),
% \]
% where \(V_\pi(s_t)\) is from trading agent and \(\pi^\alpha\) controls market dynamics through the generator. While the worst-case adversary can be optimized via any RL method, we train it via Q-learning.

As for our robust trading agent, its goal is to learn a \emph{Robust Perfect Bayesian Equilibrium}, which maximize the value function according to its current belief $\hat{\mathbf M}_{t, t-L}$ over the true macroeconomic:
\[
(\pi^*(\cdot|s_t, \hat{\mathbf M}_{t, t-L}), \pi^\alpha(\mathbf M^{\alpha, *}_{t-L}| S_{t-L})) \in \argmax_{\pi} \min_{\pi^\alpha} V_\pi(s_t,b_t).
\]
\subsubsection{Quantile Belief Network}

% To handle evolving and often adversarial market conditions, the agent maintains a belief distribution over hidden states. 

% Rather than relying on a single static estimate, it continually updates its posterior as new evidence becomes available. In an adversarial setting, these frequent Bayesian updates help prevent the exploitation of outdated or incomplete beliefs.

To handle evolving and often adversarial market conditions, the agent maintains a belief distribution over hidden states. We implement a Quantile Belief Network (QBN) to approximate this posterior distribution. For each observation 
$s$, the QBN jointly embeds both feature and temporal inputs, then passes them through an LSTM-based encoder. A normalization layer stabilizes the hidden representation, and a linear decoder outputs multiple quantiles of the market state: $b \;=\; \mathrm{QBN}(s)$. By predicting quantiles rather than fitting a single parametric form, the QBN can capture the heavy-tailed and skewed return distributions commonly observed in financial data. In practice, selecting an inappropriate target leads QBN training to non-convergency, particularly in trading environments. To mitigate this, our QBN compares its predicted return distributions to a short-term moving average (5-day window), a strong market-state indicator. Systematic deviations from this baseline may signal regime changes or adversarial behavior, prompting a more cautious trading stance. In this way, the quantile-based representation operates as a continuous monitor of partial observability and evolving market conditions.

\begin{figure*}[htbp]
    \centering
    
    % Reduce vertical space above the figure
    % \vspace{-1em}
    
    % Optionally use \scalebox to uniformly scale all subfigures
    % \scalebox{0.95}{%
    
    \begin{subfigure}[b]{0.28\textwidth}
        \includegraphics[width=\textwidth]{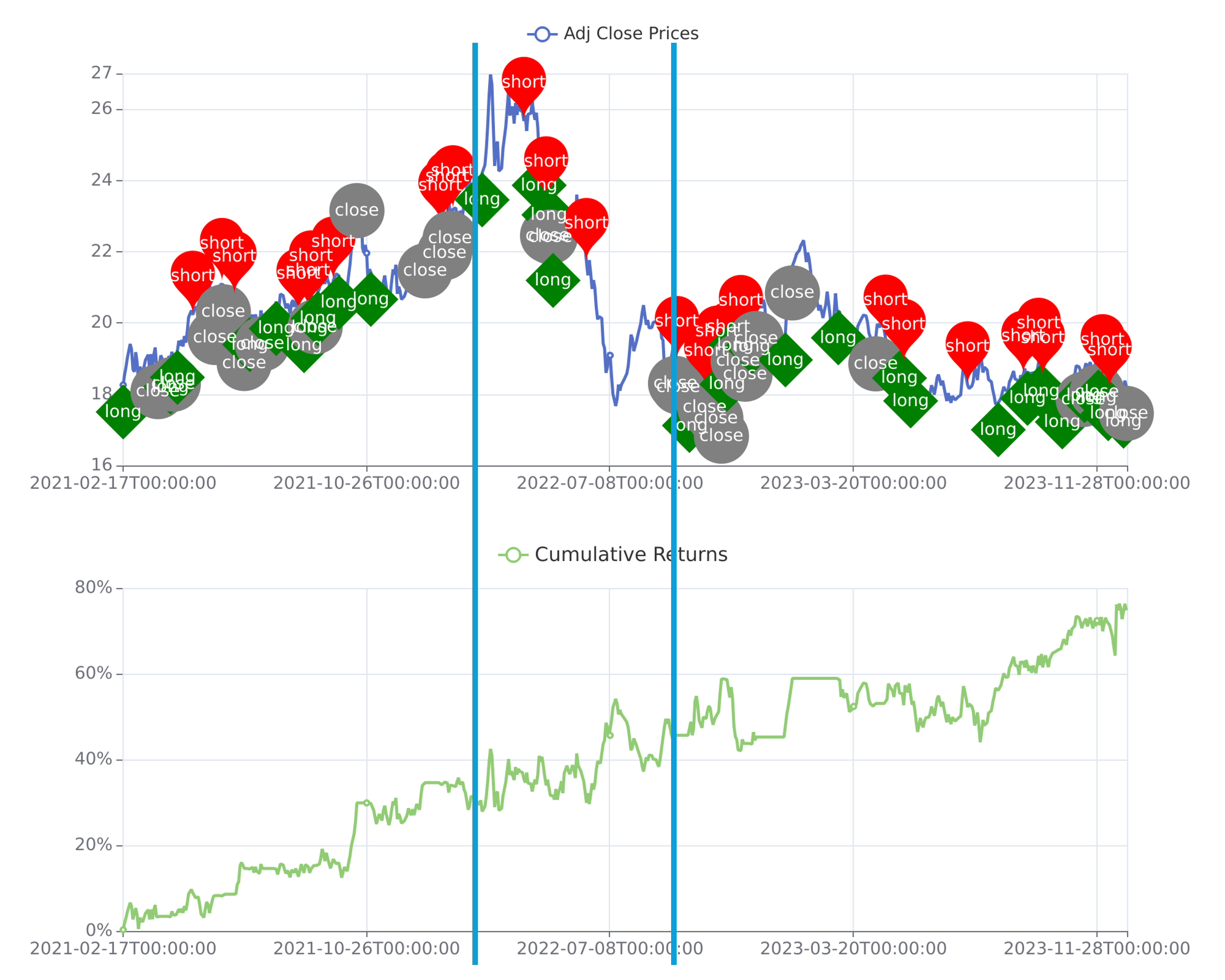}
        \caption{Our method}
        \label{fig:our_method}
    \end{subfigure}%
    % \hfill
    \begin{subfigure}[b]{0.28\textwidth}
        \includegraphics[width=\textwidth]{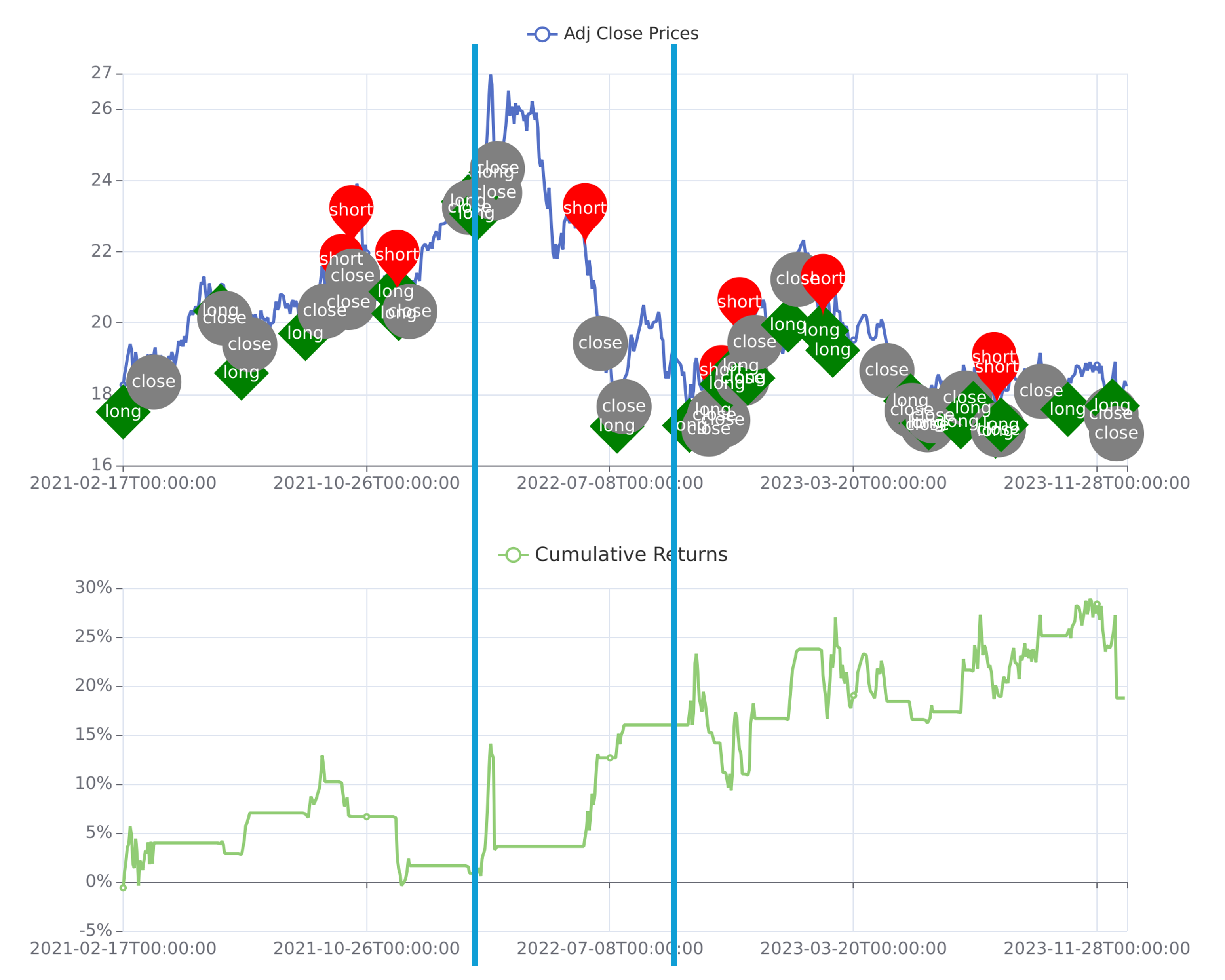}
        \caption{RARL}
        \label{fig:rarl_baseline}
    \end{subfigure}%
    % \hfill
    \begin{subfigure}[b]{0.28\textwidth}
        \includegraphics[width=\textwidth]{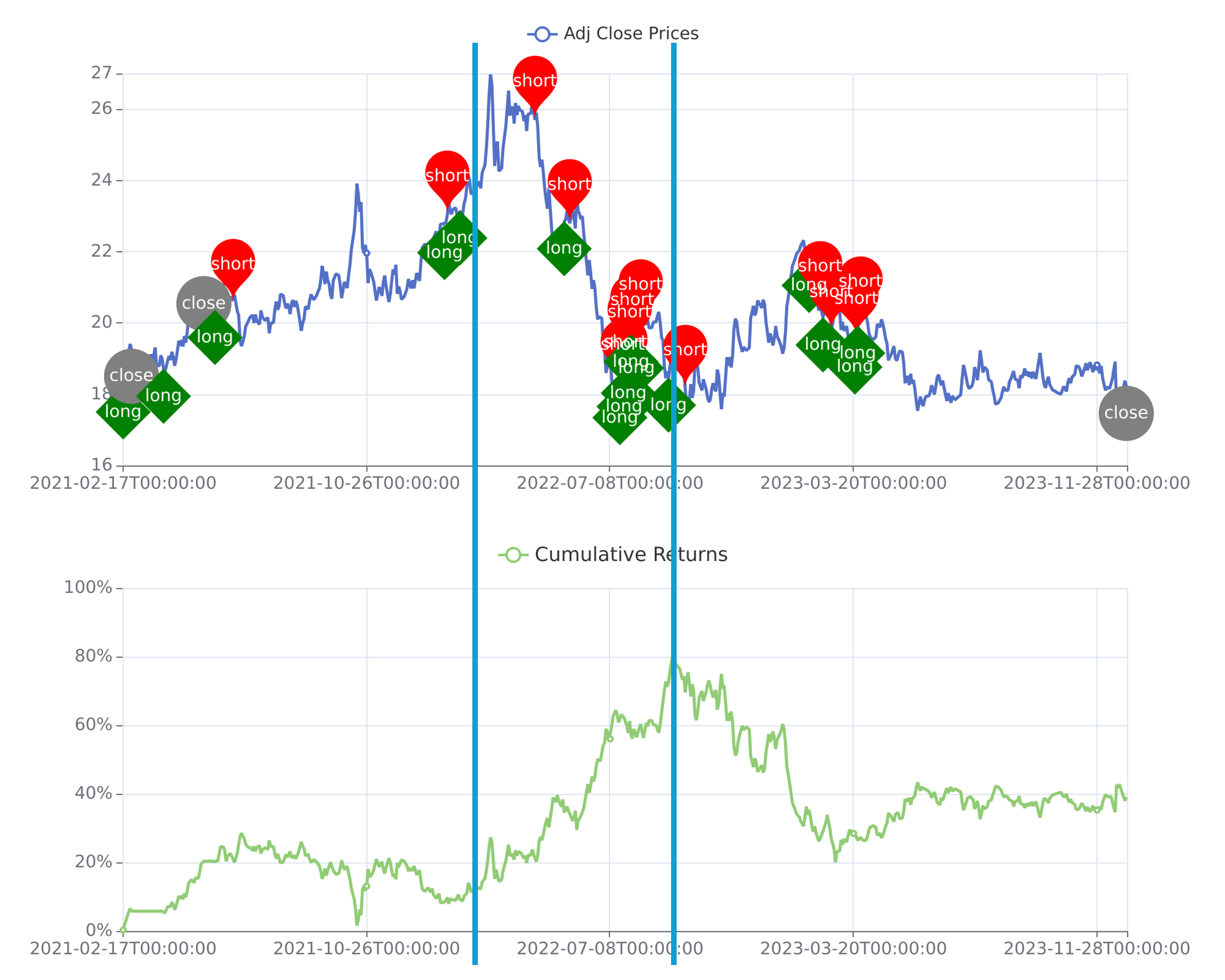}
        \caption{DQN}
        \label{fig:dqn_method}
    \end{subfigure}

    % Close the \scalebox if used
    % }
    
    % Reduce vertical space between subfigures and caption
    \vspace{-1em}
    
    \caption{Comparison of trading actions (top) and cumulative returns (bottom) for our method, RARL, and DQN on DBB from 2021 to 2024. Our method adapts to both high-volatility \textcolor{cyan!70}{pandemic phase} (between the blue lines) and calmer phases (otherwise).}
    \label{fig:case_study}
    
    % Reduce vertical space below the figure
    \vspace{-1em}
\end{figure*}

\subsubsection{Robust Trading via Bayesian Neural Fictitious Self-play}

Our trading agent seeks to maximize its cumulative reward under uncertainties, which is difficult for two reasons. First, the worst-case adversary adapts while the trading agent updates, making the environment non-stationary. Second, the agent cannot directly observe changes in these factors, so it must act optimally with its belief.

We address these challenges via \emph{Bayesian Neural Fictitious Self-Play} (Bayesian NFSP). We use Bayesian methods to learn a policy and a value function conditioned on the agent’s belief. The adversary, the trading agent, and the value function are jointly updated using NFSP for stability. We first discuss the value function conditioned on its belief $b$, which is updated by combining the Harsanyi transform \cite{harsanyi1967games} with the Bellman equation in our trading MDP:

% While the goal of our trading agent is to maximize its reward under uncertain macroeconomic factors and environments, which is challenging due to two aspects. First, the macroeconomic factors is unknown and change consistently, making the environment non-stationary. Moreover, as the policy of trading agent changes, the worst-case adversary that controls the macroeconomic factors also change as well. For example, after the trading agent learns to trade under current worst-case macroeconomic factors, the adversary can generate a new set of macroeconomic factors that sidesteps the current updated trading agent. Second, the change in macroeconomic factors is unknown to trading agent, making it unable to change its policy accurately. Thus, the goal of the trading agent is to act optimally conditioned on its belief on the market.

% To facilitate stable learning of our robust trading agents, we propose Bayesian neural fictitious self-play (Bayesian NSFP). We use Bayesian concept to learn an optimal policy conditioned on its belief, which involves learning a value function and a policy that conditions on current belief. Next, we update the value function, the adversary and the trading agent via neural fictitious self-play for stability. We first discuss the value function. Specifically, the value function of the agent is conditioned on its belief $b$, and is updated by combining Harsanyi transform \cite{harsanyi1967games} with Bellman equation in our trading MDP:
\[
\begin{aligned}
Q_\pi(s, a, b) = R(s, a) +  \gamma \sum_{s' \in \mathbf{S}} \sum_{m \in \mathbf{M}} G(x_t'|\mathbf M_{t,t-L}, \mathbf N, \mathbf{X}_{t-L}) \\ \pi^\alpha(\mathbf M^{\alpha, *}_{t-L}| \mathbf M_{t-L}. X_{t-L}) Q_\pi(s', a', b'),
\end{aligned}
\]
With belief updated via Bayes' rule, we use the QBN, ensuring faster and more stable convergence. This can be learned via TD loss:
\[
\mathcal L_{TD} = (Q_\pi(s, a, b) - R(s, a) - Q_\pi(s', a', b'))^2,
\]
This setup treats the adversary’s actions as part of the environment transitions. We use the same value function to update both the trading agent and the adversary. To achieve stable learning in this setting, we adopt NFSP, which avoids divergence or convergence to suboptimal equilibria. NFSP maintains time-averaged policies for both agents. Each agent updates its policy by training against the opponent’s time-averaged policy, and updates its own time-averaged policy with its current policy at each step. The complete process is given in the Algorithm \ref{alg:bayesiannfsp} in the Appendix \ref{section:Algorithms}.

\section{EXPERIMENTS}
\label{section:experiments}
\subsection{Dataset}
We conduct testing on nine ETFs covering commodities, FX pairs, and stock indices, dating back to each instrument's first trading day. For dataset splitting, we use data from 2018 to 2020 for validation, 2021 to 2024 for testing, and the remaining data for training across all datasets. Both the data generator and trading agent use the same dataset setting. We use 46 macroeconomic indicators sourced from the Federal Reserve Economic Data (FRED), covering key aspects such as economic activity and interest rates. The feature selection process is described in Algorithm \ref{alg:feature_selection} in the Appendix.

\subsection{Robust Trading Agent}
\textbf{Baselines.}
We compare our method against 9 baselines:  
\textbf{(1) Buy-and-Hold}: A rule-based baseline holding the instrument long throughout.  
\textbf{(2) DQN} \cite{mnih2015human}: Deep Q-learning.  
\textbf{(3) Robust Trading Agent} \cite{heinrich2016deep}: Uses QBN and NFSP with DQN. 
\textbf{(4) Naïve Adversarial} \cite{pattanaik2017robust}: Injects random noise as an attack based on (3).
\textbf{(5) RoM-Q} \cite{nisioti2021romq}: Minimizes the Q-value of the agent based on (3).
\textbf{(6) RARL (Ours w/o generator)} \cite{pinto2017robust}: Perturbs states using an adversarial agent controlling noise based on (3).
\textbf{(7) DeepScalper} \cite{sun2022deepscalper}: A risk-aware RL framework for trading.
\textbf{(8) EarnHFT} \cite{qin2024earnhft}: An three-stage hierarchical RL framework for trading.
\textbf{(9) CDQN-rp} \cite{zhu2022quantitative}: A CDQN-based RL method with a random target update for risk-aware trading.

\noindent
\textbf{Ablations.}
We compare our method against 3 ablations:  
\textbf{(10) IPG (Ours w/o Bayesian NFSP)}: Use Independent Policy Gradient \cite{daskalakis2020independent} instead of Bayesian NFSP in our method.
\textbf{(11) Ours w/o adv agent}: Our method without the adversarial agent and uses random noise as the generator control. Additionally, \textbf{(6) RARL} can be seen as \textbf{Ours w/o generator}.

\noindent
\textbf{Network Architectures}
\label{section:network}
The Encoder, Decoder, Forecaster and Discriminator of the generator consist of LSTM blocks. For the trading agent, the encoder uses an embedding layer followed by Transformer blocks with MLP and Tanh. The decoder is a linear layer. The Q-function is learned via the encoder-decoder pipeline.

\noindent
\textbf{Metrics.} We use metrics following \cite{sun2023trademaster,qin2024earnhft}, including a profit metric annual return rate (ARR) as $\frac{V_T-V_0}{V_0} \times \frac{C}{T}$, where $V$ is the net value, $T$ is the number of trading days, and $C = 252$, a risk-adjusted profit metrics Sharpe ratio (SR) as $\frac{\mathbb{E[\textbf{r}]}}{\sigma[\textbf{r}]}$, where $\textbf{r} = [{\frac{V_1-V_0}{V_0}}, {\frac{V_2-V_1}{V_1}}, ..., {\frac{V_T-V_{T-1}}{V_{T-1}}}]^{T}$, and a risk metric: maximum drawdown (MDD) measures the largest loss from any peak to show the worst case.

\subsubsection{Main Results}
\label{sec:result_analysis}

Table~\ref{tab:updated_results} compares our method with 9 baselines and 2 ablations on 9 instruments. Our model consistently outperforms across various asset classes, including Commodities (DBB, GLD, UNG, DBC) and Equities (SPY, QQQ, IWM), demonstrating strong adaptability. In Currency ETFs (FXY, FXB), while Buy-and-Hold suffers negative returns, our method delivers consistently positive ARR with lower MDD, showcasing its effectiveness in navigating macro-driven currency fluctuations.
The Wilcoxon signed-rank test shows that the superiority of our result is statistically significant (p < 0.05) as detailed in Appendix \ref{section:Wilcoxon_test}.

% \begin{itemize}
%     \item \textbf{Energy and Commodities (DBB, GLD, UNG, DBC).} 
%     Our method delivers consistently high ARR and SR while often achieving a smaller MDD. For instance, in UNG---well-known for extreme volatility---it achieves an ARR of 29.12\% and the lowest MDD of 49.37\%. In the metals-focused DBB and GLD, it balances high returns with moderated drawdowns, surpassing most baselines in the risk-adjusted metric.

%     \item \textbf{Equities (SPY, QQQ, IWM).} 
%     Across large-cap (SPY) and growth-heavy (QQQ) ETFs, our model outperforms on ARR and SR, while controlling MDD more effectively than purely adversarial methods. Notably, it excels in the small-cap IWM with an ARR of 17.32\% and a Sharpe ratio of 1.41, suggesting robust adaptability to diverse market structures.

%     \item \textbf{Currency ETFs (FXY, FXB).} 
%     While Buy-and-Hold returns are negative in FXY and FXB during this period, our method consistently exhibits positive ARR with significantly lower MDD. For example, in FXY, it posts an ARR of 15.14\% with a Sharpe ratio of 2.51, accompanied by only a 5.80\% MDD. This performance underscores its capacity to navigate macro-driven currency fluctuations effectively.
% \end{itemize}

% \subsubsection{Ablation Study: RARL vs. our method}
% \label{subsec:ablation_study}
% To highlight the contribution of our generator component, we compare our method against an adapted RARL setup. RARL alone introduces adversarial disturbances to enhance robustness. 

\noindent
\textbf{Ablations.} Baseline (6), (10), (11) are ablations of our studies.  \textbf{RARL (6)} can be seen as our method without generator. As shown in Table~\ref{tab:updated_results}, RARL does not have assess to the market dynamics (our generator), learning overly conservative policy that achieving good risk control (higher MDD), but gaining lower profit. \textbf{IPG (10)} highlights the importance of our Bayesian NFSP on stablizing training dynamics, while \textbf{Ours w/o adv agent (11)} shows optimizing with random noises makes the policy still prone to fluctuations, highlighting the importance of max-min optimization and using an adversarial agent.

% the importance of max-min optimization against a worst-case agent, not just random noises.

% The comparison of our method with RARL highlighted the contribution of our generator. As shown in Table~\ref{tab:updated_results}, RARL achieves favorable MDD values—indicating good risk control—but at the expense of profit. In contrast, our method outperforms RARL on all metrics. Our generator enriches adversarial training by synthesizing a wide range of market scenarios, thereby better preparing the policy to both manage risk and pursue profitable trades. Ablation studies (10) and (11) further justify the use of Bayesian NFSP and the adversarial agent.

\subsubsection{Case Study}
\label{sec:case_study}
We compare three policies---DQN, RARL, and 
Ours---on DBB (an ETF for metals) from 2021 to 2024. Figure~\ref{fig:case_study} shows 
trading decisions and returns for each method.
In 2021-2024, two phases emerged: I) 2022 COVID pandemic volatility peak, marked by the period between the blue lines, driven by rapid rate hikes, supply-chain disruptions, and inflation surges; II) a calmer phase other than the period in I) where markets stabilized and DBB is in sideways trends. We analyze the performance of each method:

\noindent
\textbf{DQN: Great in the Volatility, Weak Otherwise.}
DQN reacts to large price changes by maximizing returns through riskier decisions, often making "high-profit bets":
\begin{itemize}[leftmargin=*]
    \item \emph{2022 Outperformance}: Frequent trades in volatile markets allow DQN to capture major swings, yielding high gains.
    \item \emph{Losses in Calmer Markets}: Post-volatility, DQN misreads smaller fluctuations and accumulates losses with its low frequent "bet and wait" trading policy, leading to losses and drawdowns.
\end{itemize}

\noindent
\textbf{RARL: Conservative Under Stress, Misses Big Profits.}
RARL emphasizes worst-case risk management:
\begin{itemize}[leftmargin=*]
    \item \emph{Safe in Volatility and Calmer Markets}: RARL minimized exposure to 
    large shocks in 2022. In calmer markets, RARL does more frequent trading than QDN, thus avoiding losses. 
    \item \emph{Under-Exploitation of profit opportunities}: RARL avoided major losses by staying cautious but failed to capitalize on price moves. It performed consistently, though gains were the lowest among these three methods due to its conservative nature.
\end{itemize}

\noindent
\textbf{Ours: Combining Strengths of DQN and RARL.}
Our method exploits big moves (like DQN) without incurring excessive drawdowns when transitioning to the calmer market (like RARL):
\begin{itemize}[leftmargin=*]
    \item \emph{Capturing Spikes}: Like DQN, it enters significant 
    positions when volatility peaks, netting substantial gains.
    \item \emph{Adapting to Calmer Periods}: When the market is in the calmer phase, the policy transitions to a more conservative trading style, making frequent decisions and steady profits.
\end{itemize}

Although DBB did see swings before 2021, the post-2021 macro environment introduced policy-driven fluctuations. DQN capitalized on volatility bursts but struggled in subsequent calmer markets. RARL stayed safe in extremes but sacrificed potential gains. Our method displays strong performance in both volatile and low-volatility regimes, confirming that adapting across multiple macro-driven scenarios produce more robust and profitable outcomes.

\subsubsection{Time Consumption}
\label{section:time_consumption}
For the 9 ETFs we evaluated, training took an average of 22.22 h (600k steps in total) for each ETF. The inference time takes an average of 2.725 ms per step. This inference time is trivial compared with decision frequency (1 action per day), making it feasible for deployment.

\subsection{Generated Data Evaluation}
We evaluate our generated data by comparing it against generators, including TimeGAN \cite{yoon2019time}, RCWGAN \cite{he2022novel}, GMMN \cite{li2015generative}, CWGAN \cite{yu2019cwgan}, and RCGAN \cite{esteban2017real}. We also include ablation on the architecture of our generator using Informer \cite{zhou2021informer} and iTransformer \cite{liu2023itransformer}.

\begin{table}[hbpt]
% \small
\begin{tabular}{l p{1.5cm} p{1.5cm} p{1.5cm}}
\hline
\textbf{Method} & \textbf{Feature-macro} & \textbf{Inter-instrument} & \textbf{Inter-feature} \\
\hline
CWGAN   & 0.4542 & 0.3786 & 0.4732 \\
GMMN    & 0.3134 & 0.2807 & 0.3509\\
RCGAN   & 0.4258 & 0.3533 & 0.4415 \\
TimeGAN & 0.4643 & 0.3672 & 0.4590\\
RCWGAN  & 0.4584 & 0.3701 & 0.4625\\
\hline
Ours(LSTM) & 0.2678 & 0.2590 & \textbf{0.3235} \\
Ours(Informer) & \textbf{0.2523} & 0.2712 & 0.3342 \\
Ours(iTransformer) & 0.2819 & \textbf{0.2361} & 0.3371 \\
\hline
\end{tabular}
\caption{Comparison of Feature-macro, Inter-instrument, and Inter-feature correlation differences with real history data.}
\label{tab:combined_comparison}
\vspace{-1cm}
\end{table}

\subsubsection{Correlation Difference} 
We evaluate how well each model preserves real data correlation using three metrics, all measuring the difference in pairwise correlations between generated and real data:  
I) \textbf{Feature-macro}: Differences in correlations between market and macroeconomic variables;  
II) \textbf{Inter-instrument}: Differences in correlations among instruments;
III) \textbf{Inter-feature}: Differences in correlations among features within each instrument.  

Table~\ref{tab:combined_comparison} shows that our method achieves the smallest difference across all metrics, indicating superior alignment with real correlation structures and the highest level of financial fidelity.

\begin{table}[hbpt]
% \small
\begin{tabular}{l p{1.5cm} p{2cm} p{1.2cm}}
\hline
\textbf{Method} & \textbf{ReturnsACF} & \textbf{AbsReturnsACF} & \textbf{Leverage} \\
\hline
CWGAN   & 0.2088 & 0.1189 & 0.2883 \\
GMMN    & 0.0825 & 0.0766 & 0.2221 \\
RCGAN   & 0.1350 & 0.1105 & 0.2983  \\
TimeGAN & 0.2189 & 0.1320 & 0.2965 \\
RCWGAN  & 0.1840 & 0.1146 & 0.3033 \\
\hline
Ours(LSTM) & 0.0389 & 0.0461 & 0.1725 \\
Ours(Informer) & \textbf{0.0327} & \textbf{0.0458} & 0.1937 \\
Ours(iTransformer) & 0.0357 & 0.0488 & \textbf{0.1706} \\
\hline
\end{tabular}
\caption{Comparison of ReturnsACF, AbsReturnsACF, and Leverage differences with real history data.}
\label{tab:stylizedfact_comparison}
\vspace{-0.5cm}
\end{table}

\subsubsection{Market Stylized Facts} We assess how well each model captures essential time-series characteristics, which serve as key indicators of temporal correlations \cite{barberis2003style}: I) \textbf{ReturnsACF difference} measures the difference in the autocorrelation function of returns between real and generated data; II) \textbf{AbsReturnsACF difference} focuses on the autocorrelation of absolute returns, a primary indicator of volatility clustering; III) \textbf{Leverage} quantifies the difference in the correlation between past returns and future volatility, reflecting the asymmetry often observed in financial markets.

As shown in Table~\ref{tab:stylizedfact_comparison}, our approach exhibits significantly lower differences across all stylized-fact metrics, underscoring its effectiveness in replicating fundamental market dynamics.

% we use data from 2018 to 2020 for validation, 2021 to 2024 for testing,and the remaining data for training 

% analyzie what is happens to DBB train/validation/test

%  compare and analysis why our method is the best 

\section{CONCLUSION}
We proposed a Bayesian adversarial framework for robust algorithmic trading that combines macro-conditioned synthetic data generation and RL under adversarial conditions. A generator produces realistic market scenarios reflecting the changing market, while a two-player Bayesian Markov game—pitting an adversarial macro-perturbing agent against a trading agent—enables robust policy learning through adversarial training. Empirical results show that our approach significantly improves profitability and risk management over baselines, especially when adapting to unforeseen macroeconomic shifts. Furthermore, validation against competing generative models demonstrates the superior fidelity of our synthetic data. Overall, this scalable framework addresses both data realism and policy robustness in dynamic financial environments.

\section{ACKNOWLEDGMENTS}
This research is supported by the Joint NTU-WeBank Research Centre on Fintech, Nanyang Technological University, Singapore.

\bibliographystyle{ACM-Reference-Format}
\bibliography{mainpaper}

%%
%% If your work has an Appendix, this is the place to put it.
% \pagebreak
\appendix

\section{Wilcoxon Signed-Rank Tests}
\label{section:Wilcoxon_test}
\begin{table}[htbp]
\centering
\small
\caption{Directional Wilcoxon Signed-Rank Test on Annualized Return Rate (ARR)}
\label{tab:wilcoxon_arr}
\begin{tabular}{lc}
\toprule
Comparison & p-value \\
\midrule
Ours vs Buy and Hold        & 0.00195 \\
Ours vs DQN                 & 0.00195 \\
Ours vs Robust Trading Agent & 0.00391 \\
Ours vs Naïve Adversarial   & 0.00586 \\
Ours vs RoM-Q               & 0.00195 \\
Ours vs RARL                & 0.00977 \\
Ours vs DeepScalper         & 0.00195 \\
Ours vs EarnHFT             & 0.00195 \\
Ours vs CDQN-rp             & 0.00195 \\
Ours vs w/o adv agent       & 0.00195 \\
Ours vs IPG                 & 0.00586 \\
\bottomrule
\end{tabular}
\end{table}

\begin{table}[htbp]
\centering
\small
\caption{Directional Wilcoxon Signed-Rank Test on Sharpe Ratio (SR)}
\label{tab:wilcoxon_sr}
\begin{tabular}{lc}
\toprule
Comparison & p-value \\
\midrule
Ours vs Buy and Hold        & 0.00195 \\
Ours vs DQN                 & 0.00195 \\
Ours vs Robust Trading Agent & 0.01758 \\
Ours vs Naïve Adversarial   & 0.01367 \\
Ours vs RoM-Q               & 0.00977 \\
Ours vs RARL                & 0.00195 \\
Ours vs DeepScalper         & 0.00195 \\
Ours vs EarnHFT             & 0.00195 \\
Ours vs CDQN-rp             & 0.00195 \\
Ours vs w/o adv agent       & 0.00195 \\
Ours vs IPG                 & 0.00586 \\
\bottomrule
\end{tabular}
\end{table}

\begin{table}[h]
\centering
\small
\caption{Directional Wilcoxon Signed-Rank Test on Maximum Drawdown (MDD)}
\label{tab:wilcoxon_mdd}
\begin{tabular}{lc}
\toprule
Comparison & p-value \\
\midrule
Ours vs Buy and Hold        & 0.00195 \\
Ours vs DQN                 & 0.00977 \\
Ours vs Robust Trading Agent & 0.00391 \\
Ours vs Naïve Adversarial   & 0.00391 \\
Ours vs RoM-Q               & 0.00195 \\
Ours vs RARL                & 0.00195 \\
Ours vs DeepScalper         & 0.00391 \\
Ours vs EarnHFT             & 0.00195 \\
Ours vs CDQN-rp             & 0.00391 \\
Ours vs w/o adv agent       & 0.00391 \\
Ours vs IPG                 & 0.00586 \\
\bottomrule
\end{tabular}
\end{table}

The Wilcoxon signed-rank test on ARR and SR shows that our approach yields significantly better values than each baseline, while for MDD, our method achieves significantly lower drawdowns.

\section{Algorithms}
\label{section:Algorithms}

\subsection{NFSP with Adversarial Observations and Quantile Belief}
\begin{algorithm}[ht]
\SetAlgoLined
\DontPrintSemicolon
\caption{NFSP with Adversarial Observations and Quantile Belief}
\label{alg:nfsp}

\KwIn{$total\_num\_step$, initial networks: \texttt{network1, network2, network3}}
\KwOut{Trained NFSP agent, belief network, and adversarial agent.}

\textbf{Initialize:} \;
\Indp
$agent \gets agent\_nfsp \gets \text{network1}$\;
$belief\_network \gets \text{network2}$\;
$adv\_agent \gets \text{network3}$\;
$\text{adv\_buffer}, \text{nfsp\_buffer}, \text{buffer} \gets [], [], []$\;
\Indm

\For{$i \gets 1$ \KwTo $total\_num\_step$}{
  $obs \gets adv\_agent(obs) + obs$\; 
  \tcp*[l]{adversarial modification of observation}

  $belief \gets belief\_network(obs)$\;
  \tcp*[l]{quantile belief from belief network}

  $use\_avg\_policy \gets (\text{random().rand()} > \tau)$\;
  
  \uIf{$use\_avg\_policy$}{
    $action \gets agent\_nfsp(obs,~belief)$\;
  }\Else{
    $action \gets agent(obs,~belief)$\;
  }

  $(obs,~reward) \gets env.\text{step}(action)$\;
  
  $\text{buffer}.\text{insert}\bigl((obs,~action,~reward,~next\_obs)\bigr)$\;
  $\text{adv\_buffer}.\text{insert}\bigl((obs,~adv\_agent(obs),~reward)\bigr)$\;

  \If(\tcp*[f]{Only store in NFSP buffer if not average‐policy}){$\neg use\_avg\_policy$}{
    $\text{nfsp\_buffer}.\text{insert}\bigl((obs,~action)\bigr)$\;
  }

  \tcp{\small Update each component}
  \text{update\_adv\_agent}($adv\_agent,~adv\_buffer$)\;
  \tcp*[l]{policy gradient for adversarial obs generator}
  
  \text{update\_nfsp\_agent}($agent\_nfsp,~nfsp\_buffer$)\;
  \tcp*[l]{MSE loss for the average‐policy branch}
  
  \text{update\_quantile\_belief}($belief\_network,~buffer$)\;
  \tcp*[l]{quantile regression for belief network}
  
  update\_agent($agent,~buffer$)\;
  \tcp*[l]{DQN update for main Q‐function}
}
\end{algorithm}

Algorithm~\ref{alg:nfsp} outlines the training procedure for our robust trading framework, which combines Bayesian NFSP with adversarial observation perturbations and quantile-based belief modeling. The agent interacts with a perturbed environment where an adversarial agent modifies the observations. A belief network estimates quantile-based market beliefs, which are fed into the NFSP agent to guide action selection. The agent alternates between using its best-response policy and an average-policy branch, controlled by a mixing parameter $\tau$. Transitions are stored in dedicated buffers to update the adversarial agent, NFSP policy, belief network, and Q-function separately.

% \newpage

\subsection{Bayesian Neural Fictitious Self-Play}

\begin{algorithm}[hbp]
\caption{Bayesian Neural Fictitious Self-Play}
\label{alg:bayesiannfsp}

\KwIn{Q function of trading agent $Q(s_t, a_t, b_t)$, time-averaged policy $\overline{\pi}$, trained market simulator $G(x_t|\mathbf{m}_{t,t-L}, \mathbf{n}, \mathbf{x}_{t-L})$, circular buffer $\mathcal M_{RL}$ and reservoir buffer for time-averaged policy $\mathcal M_{SL}$.}
\KwOut{Robust trading policy $Q(s_t, a_t, b_t)$.}

\BlankLine
Initialize Q function of trading agent, initialize the network of time-averaged policy $\overline{\pi}$, circular buffer $\mathcal M_{RL}$, reservoir buffer $\mathcal M_{SL}$, and belief \( b_0 \)\;
\For{each training iteration}{
    \For{each episode}{
        \For{each timestep \( t \)}{
            Adversary update macroeconomic factors $\mathbf M^{\alpha, *}_{t-L}$, environment proceeds via $G$ \;
            Trading agent updates belief $b_t$. Samples an action from $Q(s_t, a_t, b_t)$ with probability $\eta$ with epsilon greedy exploration, sample an action from $\overline{\pi}$ with probability $1-\eta$ \;
        }
        Store transition in circular buffer $\mathcal M_{RL}$\;
        \If{Action is sampled from $Q(s_t, a_t, b_t)$}{
            Store transition in circular buffer $\mathcal M_{RL}$\;
        }
    }
    Update value function via TD loss: \[\mathcal{L}_{TD} = (Q_\pi(s_t, a_t, b_t) - r_t - \gamma Q_\pi(s_{t+1}, a_{t+1}, b_{t+1}))^2\]
    Update time-averaged trading policy $\overline{\pi}$ via supervised learning \;
}
\Return Optimized trading policy \( \pi^\theta \)\;
\end{algorithm}

Algorithm~\ref{alg:bayesiannfsp} presents the training procedure of our proposed Bayesian Neural Fictitious Self-Play (BNFSP) framework. This method extends traditional NFSP by incorporating a belief modeling component and adversarial market simulation. During training, the agent interacts with a market simulator $G$ perturbed by adversarial macroeconomic factors, and makes decisions based on both its Q-function and a time-averaged policy. A quantile-based belief $b_t$ is updated at each step to capture latent market states. Transitions are stored in two separate replay buffers: a circular buffer for reinforcement learning updates and a reservoir buffer for supervised learning of the average policy. The agent’s Q-function is optimized via temporal-difference learning, while the average policy is updated through supervised learning. This joint training improves the robustness and adaptability of the trading agent in dynamic and uncertain environments.

% \newpage
\subsection{Correlation-Weighted Imputation}
\begin{algorithm}[htbp]
\caption{Correlation-Weighted Imputation}
\label{alg:correlation_weighted_imputation}

\textbf{Input:} Ticker set \( T = \{t_1, \ldots, t_N\} \), feature datasets \( \{D_t\}_{t \in T} \), correlation matrices \( \{C^{(f)}\}_{f \in \mathcal{F}} \).\\
\textbf{Output:} Imputed datasets \( \{D_t\}_{t \in T} \).

For each feature \( f \in \mathcal{F} \), process all tickers \( t \in T \). For a given ticker \( t \), every time index \( i \) where \( D_t(i, f) \) is missing\;
Identify the set of valid tickers:
\[
V = \{ t' \in T \setminus \{t\} : D_{t'}(i, f) \text{ is available} \}
\]
\If{\( V \neq \emptyset \)}{
    Compute weights \( w_{t'} = \exp(C^{(f)}(t, t')) \) for all \( t' \in V\)\;
    Normalize weights:
    \[
    \tilde{w}_{t'} = \frac{w_{t'}}{\sum_{s \in V} w_s}
    \]
    Impute missing value:
    \[
    D_t(i, f) = \sum_{t' \in V} \tilde{w}_{t'} \cdot D_{t'}(i, f)
    \]
}

\Return Updated datasets \( \{D_t\}_{t \in T} \)\;

\end{algorithm}

To handle missing values in cross-sectional time-series data, we propose a correlation-weighted imputation method that leverages the structural similarity across assets. As detailed in Algorithm~\ref{alg:correlation_weighted_imputation}, the method imputes missing feature values for each ticker by taking a weighted average of the corresponding values from other tickers, where the weights are derived from an exponential transformation of their historical correlation. This approach ensures that more correlated tickers contribute more significantly to the imputed value, preserving consistency across assets while mitigating the noise introduced by unrelated instruments.

% \newpage

\subsection{t-SNE Plot Generation}
\begin{algorithm}[htbp]
\caption{t-SNE Plot Generation}
\label{alg:tsnegeneration}
\textbf{Input:} Raw OHLCV time series data

\textbf{Parameters:} Window size $w = 21$, t-SNE output dimension $d = 1$, perplexity $= 50$, iterations $= 3000$

\begin{itemize}
  \item[\textbf{Step 1}] Compute derived features from raw OHLCV, such as:
    \begin{itemize}
        \item Rolling returns
        \item Moving averages
        \item Standard deviations
        \item Volume-based indicators
    \end{itemize}
    Result: feature matrix $F$

  \item[\textbf{Step 2}] Construct sliding windows:
    \begin{itemize}
        \item For each time $t = w$ to $T$:
        \begin{itemize}
            \item Flatten the window $F_{t-w+1:t}$ into a vector $X_t$
            \item Assign corresponding target value $Y_t$
        \end{itemize}
    \end{itemize}
    Result: windowed feature vectors $X$, aligned targets $Y$

  \item[\textbf{Step 3}] Apply t-SNE separately:
    \begin{itemize}
        \item Compute $Z_x \leftarrow \text{t-SNE}(X, d=1)$
        \item Compute $Z_y \leftarrow \text{t-SNE}(Y, d=1)$
    \end{itemize}
    Result: 1D embeddings for features and targets

  \item[\textbf{Step 4}] Plot:
    \begin{itemize}
        \item Use $Z_x$ as x-axis and $Z_y$ as y-axis
        \item Color each point according to chronological order (e.g., train vs. test)
    \end{itemize}
\end{itemize}
\end{algorithm}

To visualize the distributional shift between training and testing states, we construct a 1D embedding of both state features and reward targets using t-SNE. As described in Algorithm~\ref{alg:tsnegeneration}, we first extract meaningful technical features from raw OHLCV data and organize them into overlapping windows. Each window is flattened into a vector representation, aligned with its corresponding future return. We then apply t-SNE separately to the feature and target spaces to obtain low-dimensional embeddings that preserve local structure. By plotting these embeddings against each other and coloring by temporal phase, we can effectively illustrate the dynamics shift faced by the agent between training and testing periods.

% \newpage
\subsection{Feature Selection}

\begin{algorithm}[htbp]
\caption{Feature Selection Procedure}
\label{alg:feature_selection}
\textbf{Input:} \\
\hspace{1em} Raw feature set $\mathcal{F} = \{f_1, f_2, ..., f_n\}$ \\
\hspace{1em} Future return series $\mathbf{r}_{t+\Delta}$ \\
\hspace{1em} Correlation threshold $\tau_{\text{corr}}$ \\
\hspace{1em} Redundancy threshold $\tau_{\text{red}}$ \\

\vspace{0.5em}
\textbf{Output:} \\
\hspace{1em} Selected feature set $\mathcal{F}_{\text{selected}}$ \\

\vspace{0.5em}
\textbf{Procedure:}
\begin{enumerate}
    \item Initialize $\mathcal{F}_{\text{candidate}} \leftarrow \emptyset$
    \item For each feature $f_i$ in $\mathcal{F}$:
    \begin{itemize}
        \item Compute Pearson correlation $\rho_i = \text{corr}(f_i, \mathbf{r}_{t+\Delta})$
        \item If $|\rho_i| > \tau_{\text{corr}}$, add $f_i$ to $\mathcal{F}_{\text{candidate}}$
    \end{itemize}
    \item Initialize $\mathcal{F}_{\text{selected}} \leftarrow \emptyset$
    \item For each $f_i$ in $\mathcal{F}_{\text{candidate}}$:
    \begin{itemize}
        \item Compute correlation $\rho_{ij} = \text{corr}(f_i, f_j)$ for all $f_j \in \mathcal{F}_{\text{selected}}$
        \item If $\max_j |\rho_{ij}| < \tau_{\text{red}}$, add $f_i$ to $\mathcal{F}_{\text{selected}}$
    \end{itemize}
    \item Return $\mathcal{F}_{\text{selected}}$
\end{enumerate}
\end{algorithm}

To ensure that the input features are both informative and non-redundant, we employ a two-stage correlation-based feature selection procedure, as detailed in Algorithm~\ref{alg:feature_selection}. In the first stage, we compute the Pearson correlation between each candidate feature and the target future return, retaining only those with correlation magnitude above a threshold $\tau_{\text{corr}}$. In the second stage, we iteratively filter out redundant features by enforcing a maximum pairwise correlation constraint $\tau_{\text{red}}$ with already selected features. This procedure results in a compact feature set that preserves predictive relevance while mitigating multicollinearity.

\end{document}